%% file: bare_jrnl_compsoc.tex
\documentclass[10pt,journal,compsoc]{IEEEtran}
\usepackage[nocompress]{cite}

\ifCLASSINFOpdf
\else
\fi
\usepackage{amsmath}
\usepackage{amssymb}
\usepackage{epsfig}
\usepackage{graphicx}
\usepackage{multirow}
\usepackage{threeparttable}
\usepackage{comment}
\usepackage[tight]{subfigure}
\usepackage{caption}
\usepackage{subfig}
\usepackage{csquotes}
\usepackage{algorithm}
\usepackage{multirow}
\usepackage{color}
\usepackage[noend]{algpseudocode}

\newcommand{\tabincell}[2]{\begin{tabular}{@{}#1@{}}#2\end{tabular}}%

\usepackage{array}

\newcolumntype{P}[1]{>{\centering\arraybackslash}p{#1}}
\newcolumntype{M}[1]{>{\centering\arraybackslash}m{#1}}

\begin{document}
	
	\setlength{\abovedisplayskip}{5pt}
	\setlength{\belowdisplayskip}{5pt}
	\setlength{\textfloatsep}{15pt}
%
\title{Automated Latent Fingerprint Recognition}
%
%
%
%

\author{Kai~Cao and~Anil~K.~Jain,~\IEEEmembership{Fellow,~IEEE}

\IEEEcompsocitemizethanks{

\IEEEcompsocthanksitem Kai Cao and A.K. Jain are with the Dept. of Computer Science and Engineering, Michigan State University, East Lansing, MI 48824 U.S.A. \protect\\
E-mail: \{kaicao,jain\}@cse.msu.edu

}
\thanks{}
}

\date{\today}

%
%

\markboth{}%
{}
%


\IEEEcompsoctitleabstractindextext{%
\begin{abstract}
Latent fingerprints are one of the most important and widely used evidence in law enforcement and forensic agencies worldwide. 
Yet, NIST evaluations show that the performance of state-of-the-art latent recognition systems is far from satisfactory. An automated latent fingerprint recognition system with high accuracy is essential to compare latents found at crime scenes to a large collection of reference prints to generate a candidate list of possible mates.  In this paper, we propose an automated latent fingerprint recognition algorithm that utilizes Convolutional Neural Networks (ConvNets) for ridge flow estimation and minutiae descriptor extraction, and extract complementary templates (two minutiae templates and one texture template) to represent the latent.
The comparison scores between the latent and a reference print based on the three templates are fused to retrieve a short candidate list from the reference database. Experimental results show that
the rank-1 identification accuracies (query latent is matched with its true mate in the reference database) are 64.7\% for the NIST SD27 and 75.3\% for the WVU latent databases, against a reference database of 100K rolled prints. These results are the best among published papers on latent recognition and competitive with the performance (66.7\% and 70.8\% rank-1 accuracies on NIST SD27 and WVU DB, respectively) of a leading COTS latent Automated Fingerprint Identification System (AFIS). By score-level (rank-level) fusion of our system with the commercial off-the-shelf (COTS) latent AFIS, the overall rank-1 identification performance can be improved from 64.7\% and 75.3\%  to 73.3\% (74.4\%) and 76.6\% (78.4\%) on NIST SD27 and WVU latent databases, respectively.

\end{abstract}


}

\maketitle

\IEEEdisplaynotcompsoctitleabstractindextext

%
\IEEEpeerreviewmaketitle

\section{Introduction}

\begin{figure}
	\begin{center}
		\includegraphics[width=0.95\linewidth]{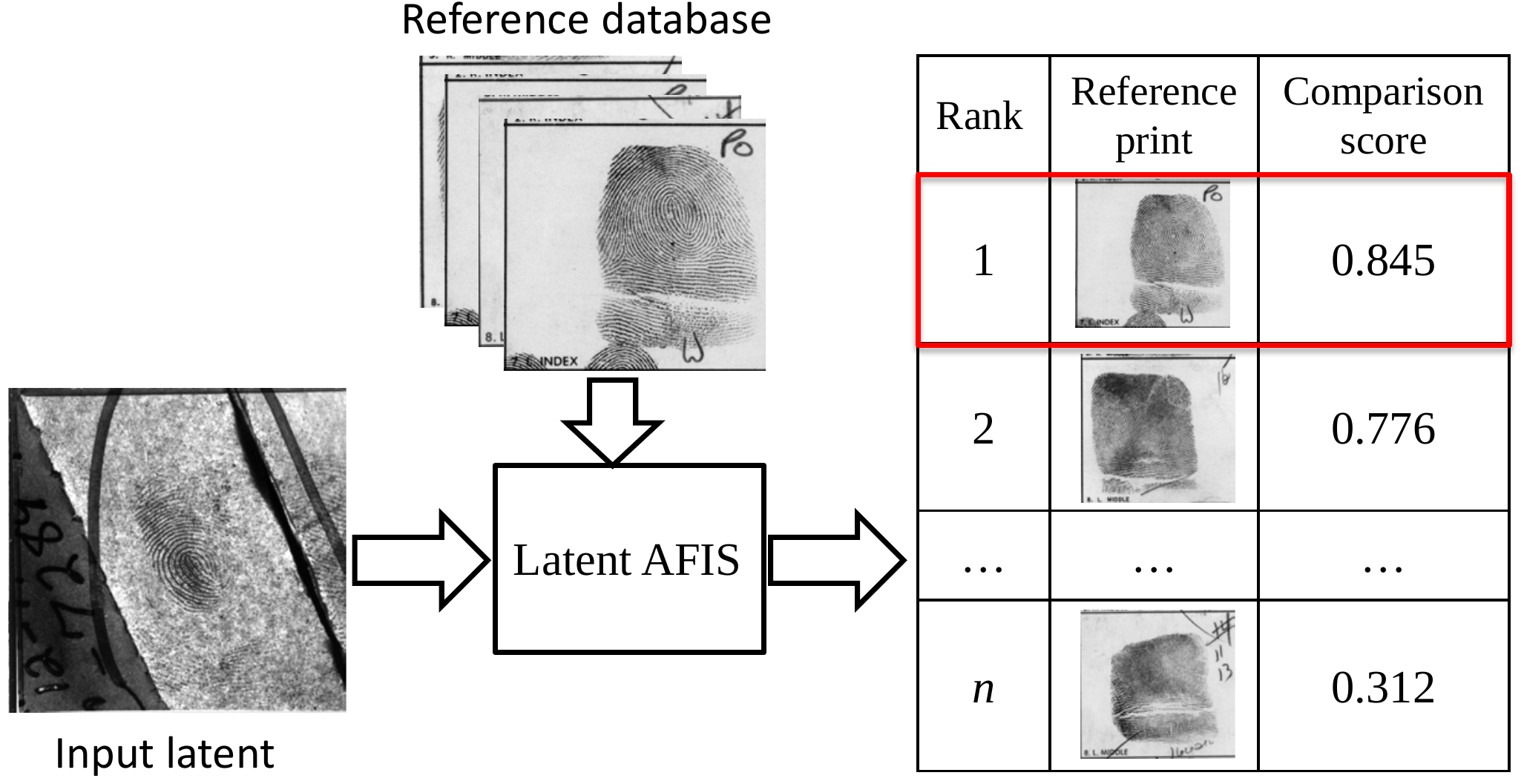}
	\end{center}
	\caption{Automated latent recognition framework. A latent image is input to a latent AFIS, and the top $n$ candidates with their comparison scores are presented to a latent expert. The  number of candidates, $n$, examined is typically less than 20. The true mate in this example is outlined in red. 
	}
	\label{fig:LatentSearch}
\end{figure}

\begin{figure}[h]	
	\begin{center}
			\subfigure[]{
			\includegraphics[width=0.4\linewidth]{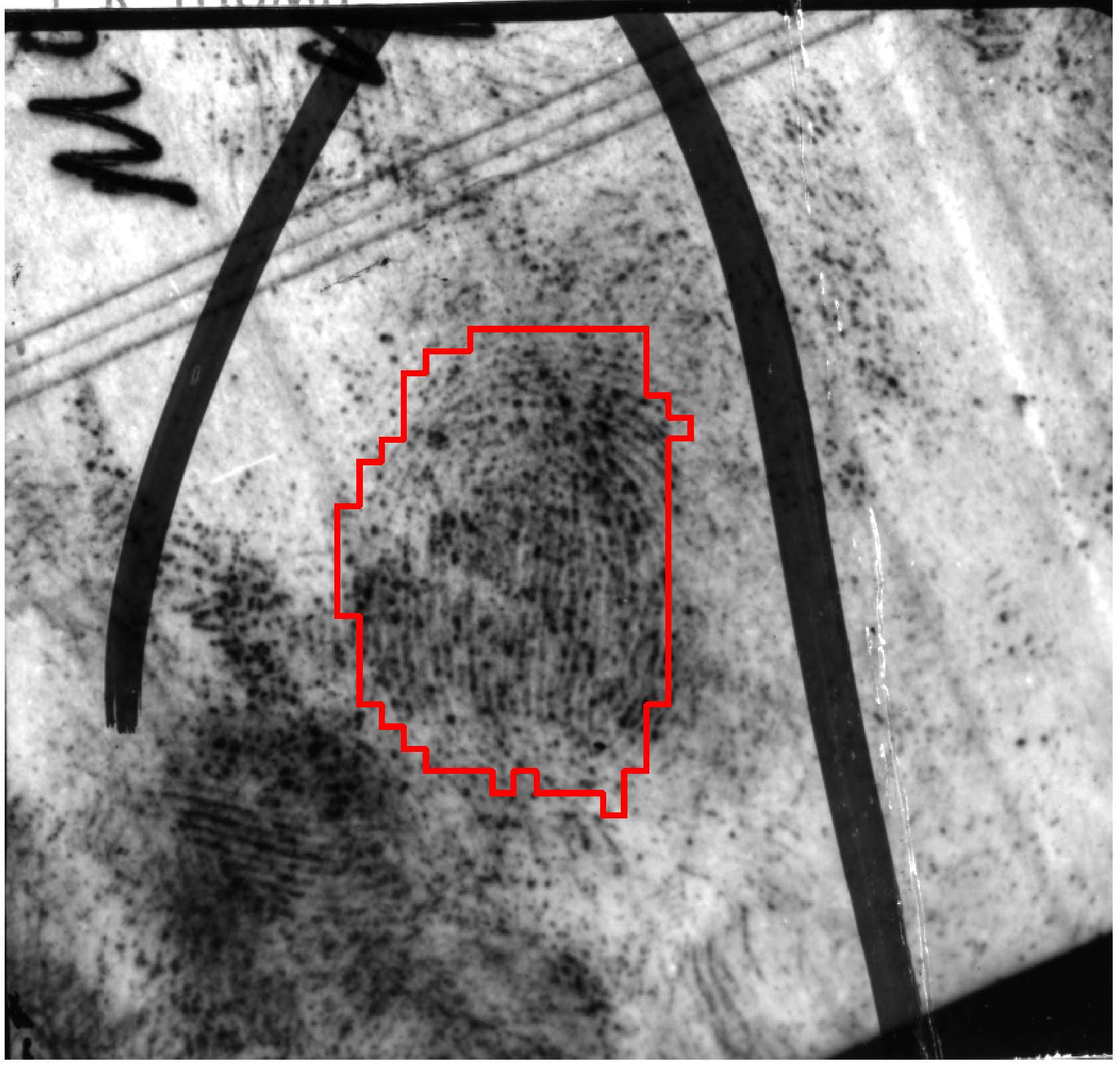}
		} \hspace{0.1cm}
		\subfigure[]{
			\includegraphics[width=0.35\linewidth]{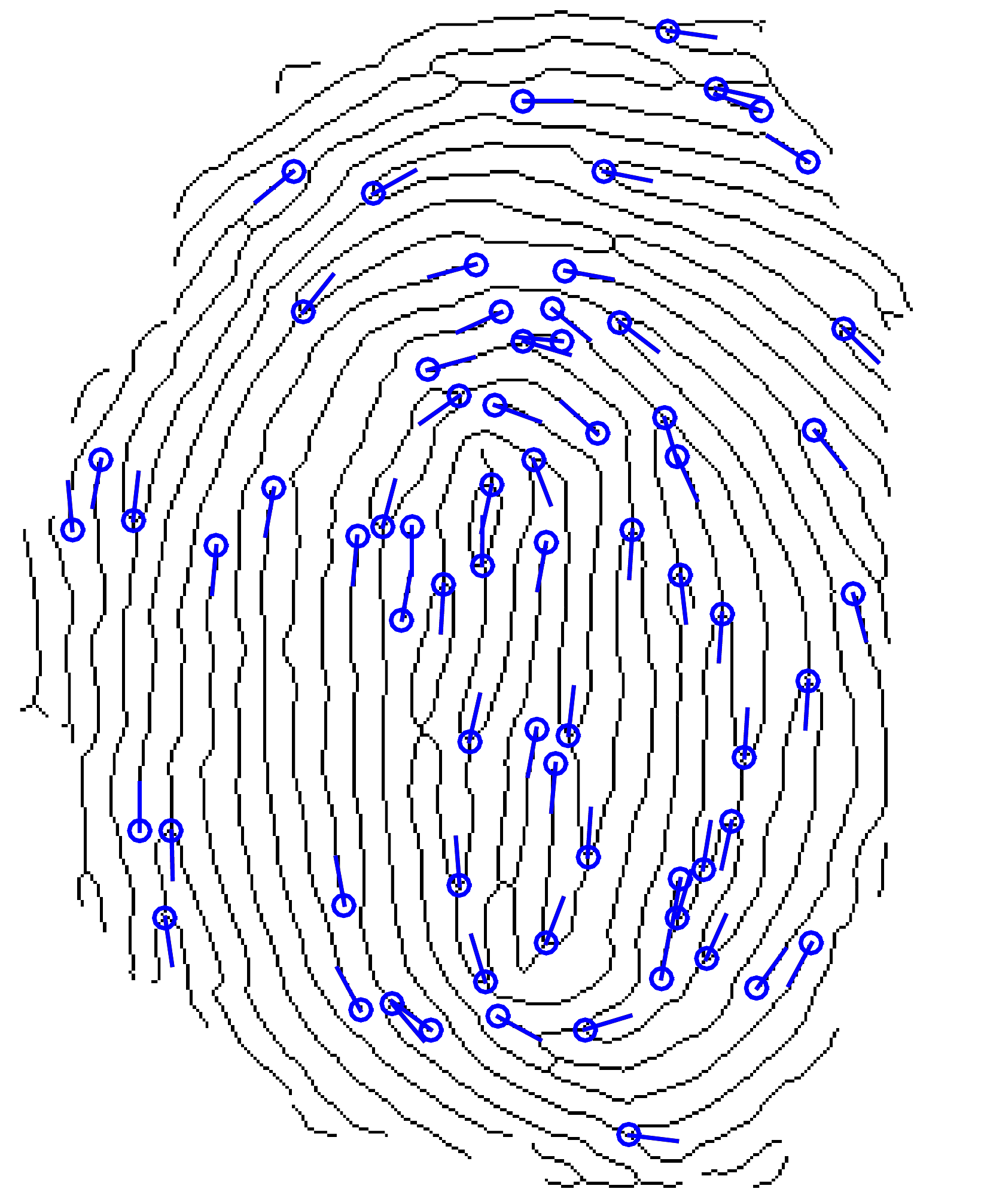}
		}
		\subfigure[]{
			\includegraphics[width=0.4\linewidth]{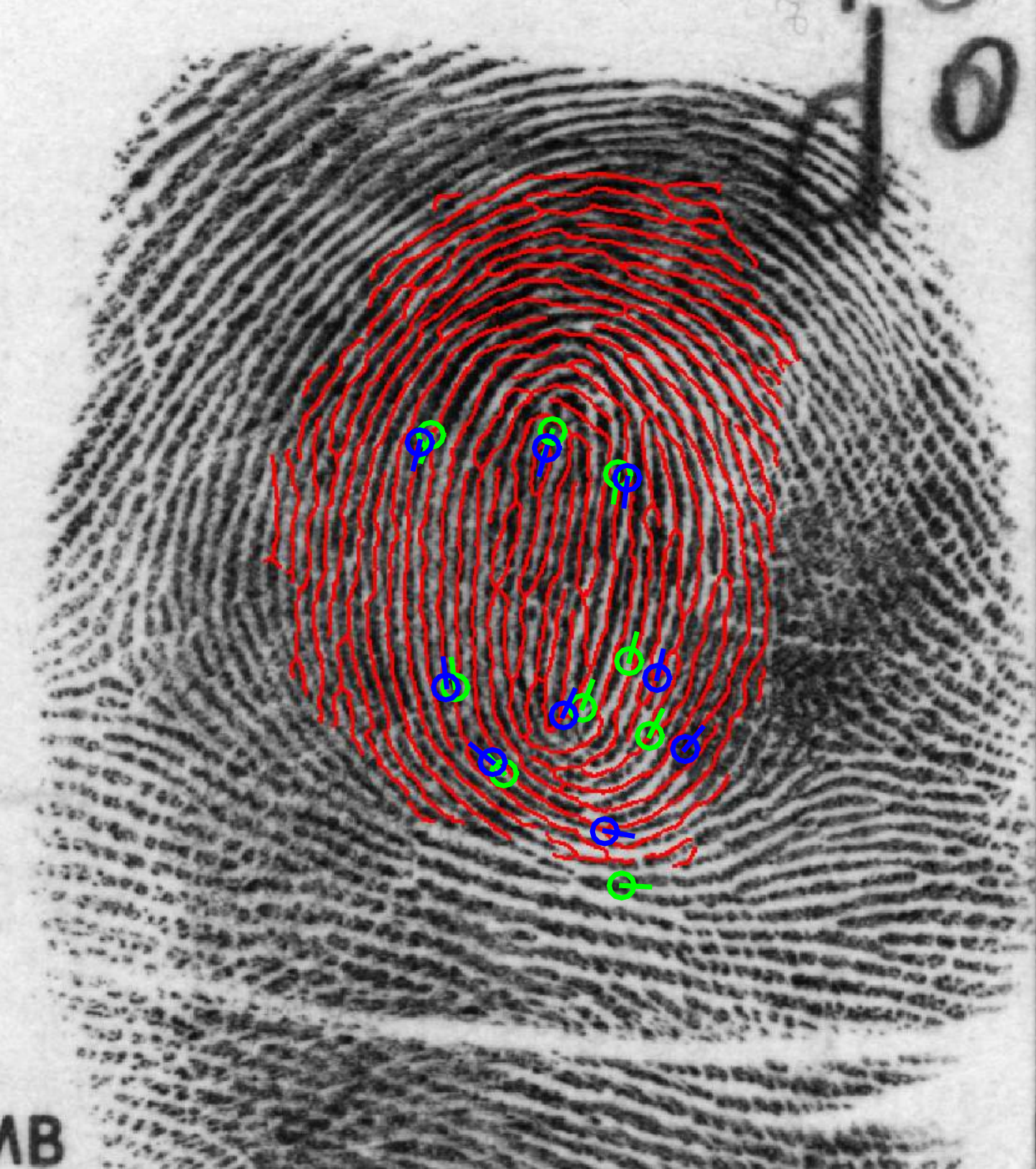}
		} \hspace{0.1cm}
		\subfigure[]{
			\includegraphics[width=0.4\linewidth]{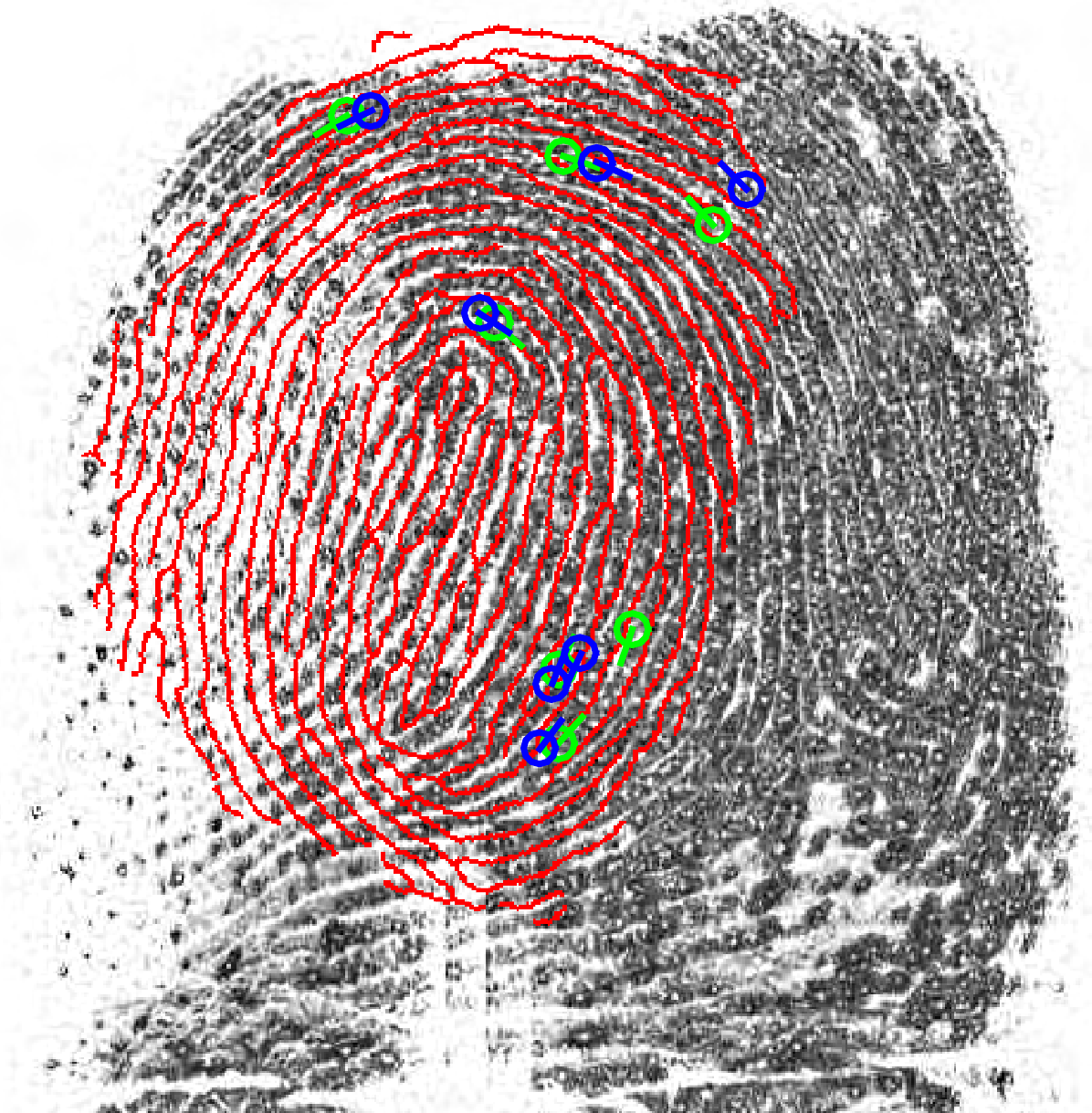}
		}
	\end{center}
	\caption{ Illustration of latent to reference (rolled) comparison. (a) Input latent with ROI outlined in red, (b) automatically extracted minutiae in (a) shown on the latent skeleton, (c) alignment and minutiae correspondences between the latent  and its true mate (rank-1 retrieval) and (d) alignment and  minutiae correspondences between the latent and the rank-2 retrieved rolled print. Blue circles denote latent minutiae and green circles denote rolled minutiae.} 	
	\label{fig:Alignment}
\end{figure}

\IEEEPARstart{E}{ver} since latent fingerprints  (latents or marks\footnote{Latent and mark both refer to a partial and smudgy friction ridge impression from an unknown source.
The term latent is preferred in North America while mark is preferred outside North America \cite{Ulery2013}. We adopt the term latent here to be consistent with our previous work \cite{Feng2011PAMI}, \cite{CaoPAMI2014}.}) were first introduced as evidence to convict a suspect in Argentina in 1893,  they have become one of the most important and widely used sources of evidence in law enforcement and forensic agencies worldwide \cite{Hawthorne2002}.  Latent fingerprint recognition requires recognizing the mate of a latent print evidence in a database of reference prints (rolled or slap fingerprints). See Figs. \ref{fig:LatentSearch} and \ref{fig:Alignment}. 
A majority (60\%) of crime laboratories in the United States reported analyzing latent fingerprints recovered from crime scenes, and a total of 271,000 latent prints were processed by public forensic crime laboratories in 2009 alone\footnote{Bureau of Justice Statistics, Census of Publicly Funded Forensic Crime Laboratories, 2009.}. During January 2017, FBI's Integrated Automated Fingerprint Identification
System (IAFIS), which maintains the largest criminal fingerprint database in the world, conducted 17,758 latent ``feature" searches (latent features were manually marked by latent examiners), and an additional 4,160 latent ``image" searches \cite{factsheep} (latent features were automatically extracted by IAFIS). 

Compared to rolled and slap prints (or reference prints), which are acquired  under supervision, latent prints are lifted after being unintentionally deposited by a subject, e.g., at crime scenes, typically resulting in poor quality in terms of ridge clarity and large background noise. Unlike reference prints, the action of depositing finger mark on a surface is not repeatable if latent prints are found to be of poor quality. National Institute of Standards \& Technology (NIST) periodically conducts technology evaluations of fingerprint recognition algorithms, both for rolled (or slap) and latent prints.  In NIST's most recent evaluation of rolled and slap prints, FpVTE 2012, the best performing Automated Fingerprint Identification System (AFIS)
achieved a false negative identification rate (FNIR) of 1.9\% for single index fingers, at a false positive identification rate (FPIR) of 0.1\% using 30,000 search subjects (10,000 subjects with mates and 20,000 subjects with no mates) \cite{FpVTE2012}. For latent prints, the most recent evaluation is the NIST ELFT-EFS where the best performing automated latent recognition system could only achieve a rank-1
identification rate of 67.2$\%$ in searching 1,114 latents against
a background containing 100,000 reference prints
\cite{Indovina2012}. The rank-1 identification rate of the best performing latent
AFIS was improved from 67.2\% to 70.2\%\footnote{The best result using both markups and images is 71.4\% rank-1 accuracy.} \cite{Indovina2012} when
feature markup by a latent expert was also input, in
addition to the latent images, to the AFIS.
This gap between reference fingerprint recognition and latent fingerprint recognition capabilities is primarily due to the poor quality of friction ridges in latent prints. This underscores the need for developing automated latent recognition with high accuracy\footnote{In forensics and law enforcement, automated latent recognition is also referred as \textit{lights-out recognition} where the objective is to minimize the role of latent examiners in latent recognition. }. 


\subsection{Current Practice}

The standard procedure for latent recognition, as practiced in forensics agencies, involves four phases: \textit{ Analysis}, \textit{Comparison}, \textit{Evaluation}, and \textit{Verification} (ACE-V) \cite{Ashbaugh1999}.
A number of studies have highlighted limitations of the ACE-V methodology.
\begin{enumerate}
\item \emph{Repeatability/reproducibility of feature markup.} Ulery et al.  \cite{Ulery2016} and Arora et al. \cite{Arora2015ICB} observed  a
    large variation among the feature markups on the same latent provided by different examiners  which affects the latent recognition accuracy \cite{Indovina2012}. The median value of markup reproducibility was found to be  only 46\% \cite{Ulery2016}. 

\item \emph{Repeatability/reproducibility of decision.} Examiner repeatability of comparison decisions  was found to be 90.0\% for mated pairs, and only 85.9\% for non-mated pairs \cite{ulery2012repeatability}. These values were even lower for comparisons assessed by the examiners as ``difficult" (i.e., low quality latents).

\item \emph{Throughput.} Manual markup requires significant effort ($\sim$15 min/latent\footnote{https://www.noexperiencenecessarybook.com/eVbzD/microsoft-powerpoint-nist-fingerprint-testing-standards-v2-02282013-pptx.html}) by latent examiners. 

\item \emph{Bias.} Since the second examiner in the verification phase is only assessing the  comparison decision made by the first examiner,
 it creates the potential for confirmation bias (see page 90 in \cite{PCAST}). 
      
\end{enumerate}

Given the current state of latent processing that relies heavily on forensic examiners, an automated latent recognition algorithm is urgently needed to give an accurate,  reliable and efficient (i.e., a short) candidate list for ever growing case workload.
An automated latent recognition system will also assist in developing quantitative validity and reliability measures\footnote{AFIS available from vendors neither provide the latent features they extract nor the true comparison scores between a latent and a reference print.} for latent fingerprint evidence as highlighted in the 2016 PCAST \cite{PCAST} and  the 2009 NRC \cite{PathForward} reports. 


\begin{figure}[t]

\begin{center}
\subfigure[]{
 \includegraphics[width=0.4\linewidth]{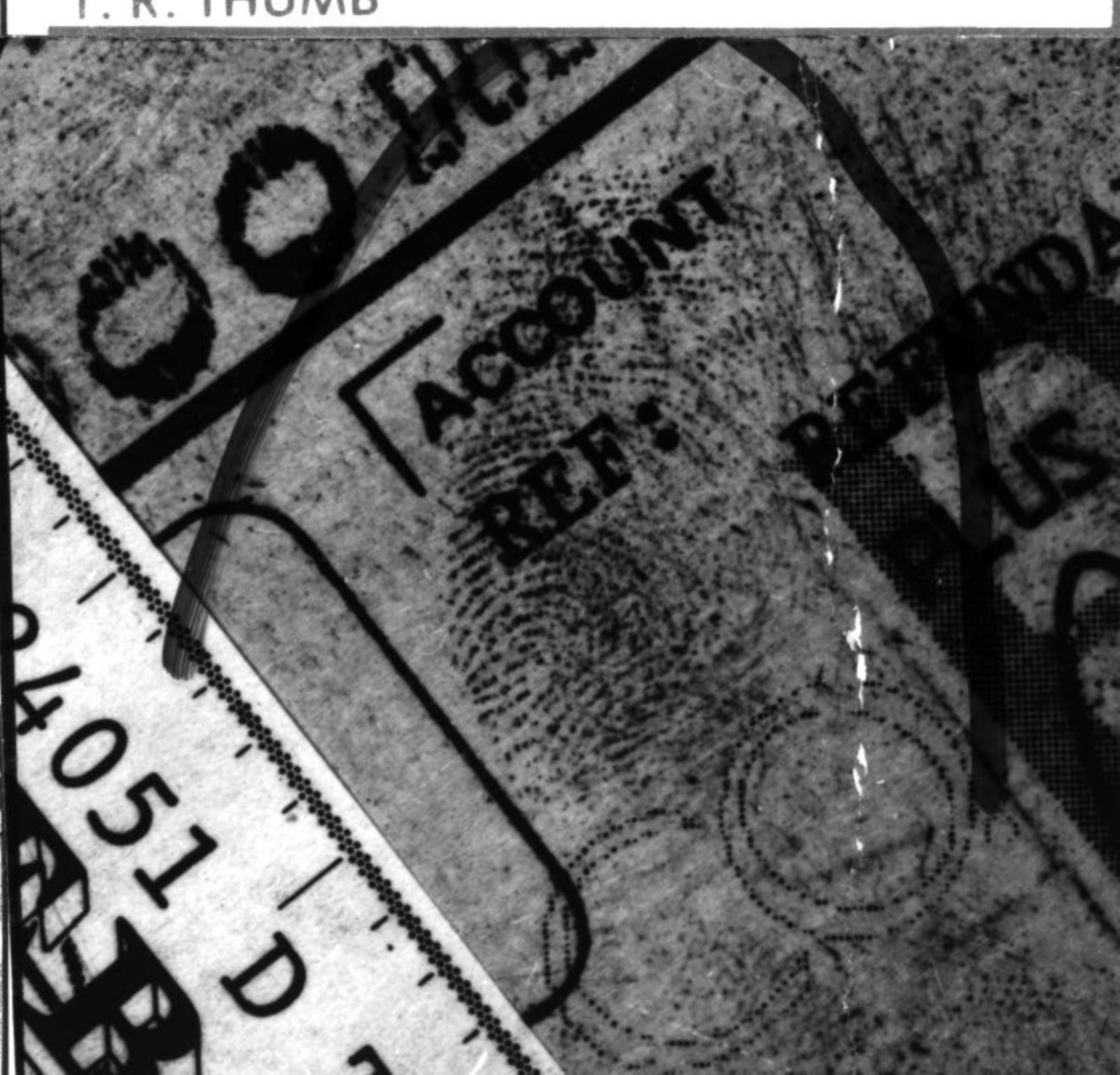}
 }\hspace{0.4cm}
 \subfigure[]{
 \includegraphics[width=0.4\linewidth]{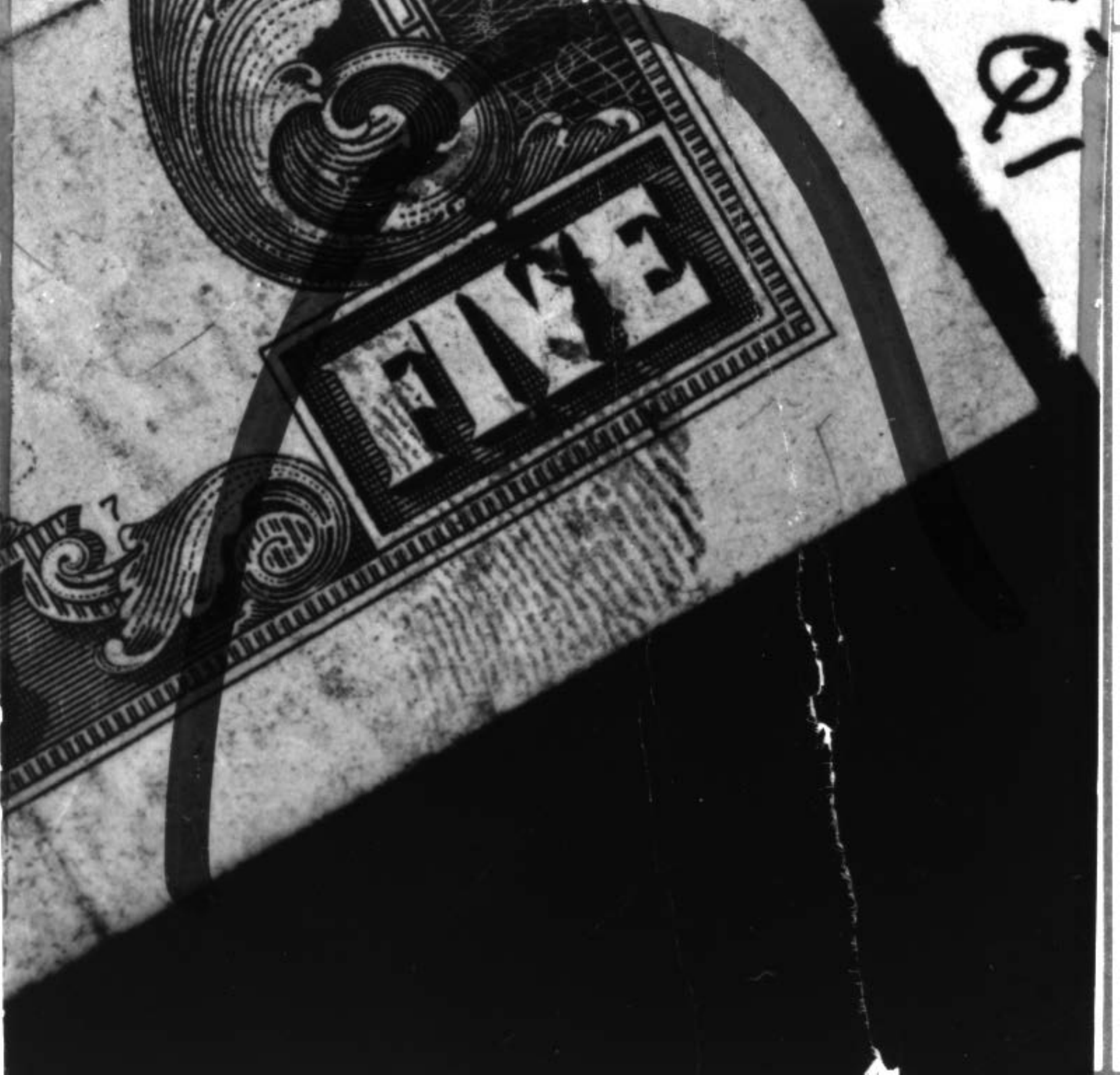}
 }
\subfigure[]{
	\includegraphics[width=0.4\linewidth]{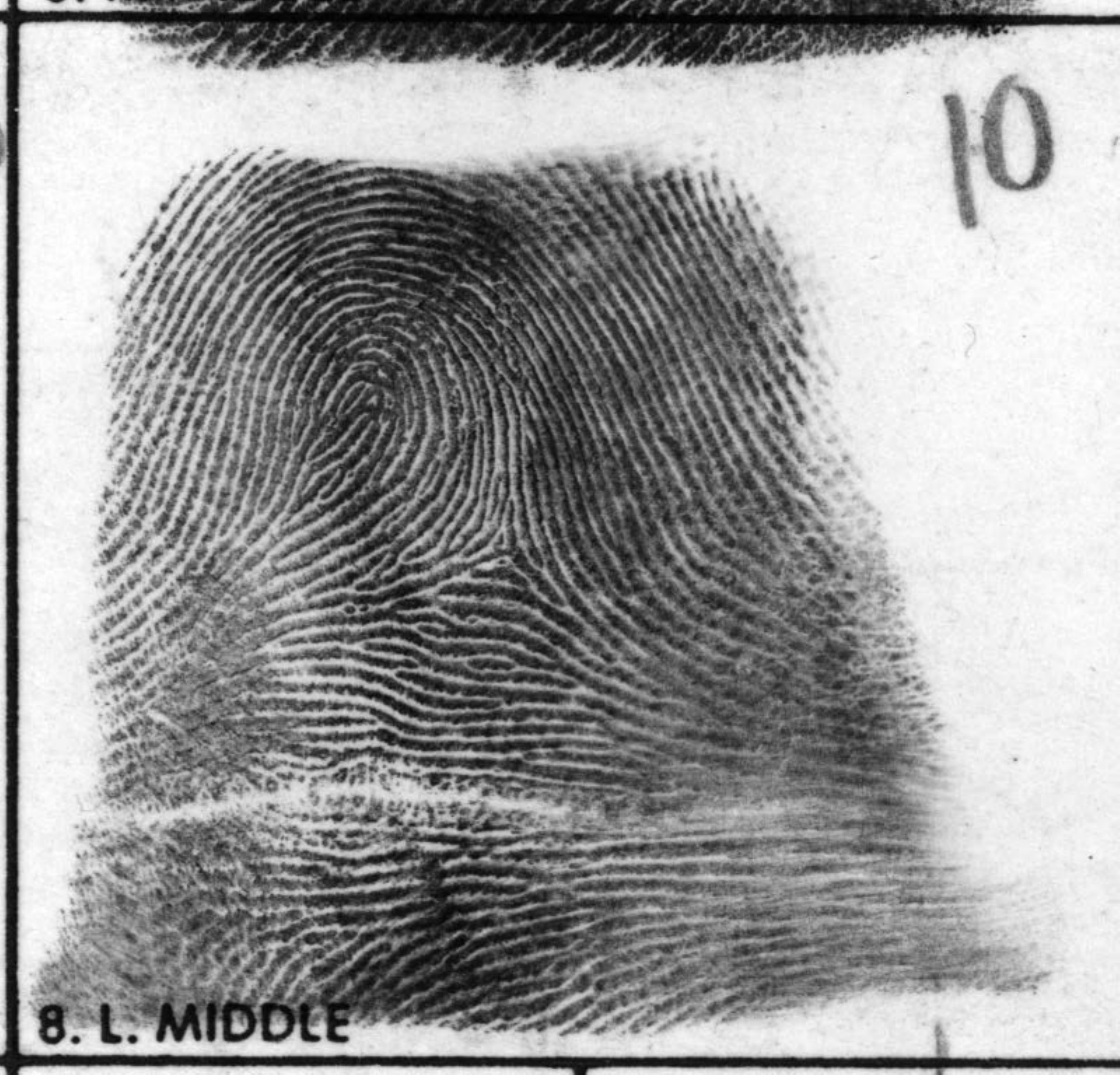}
}\hspace{0.4cm}
\subfigure[]{
	\includegraphics[width=0.4\linewidth]{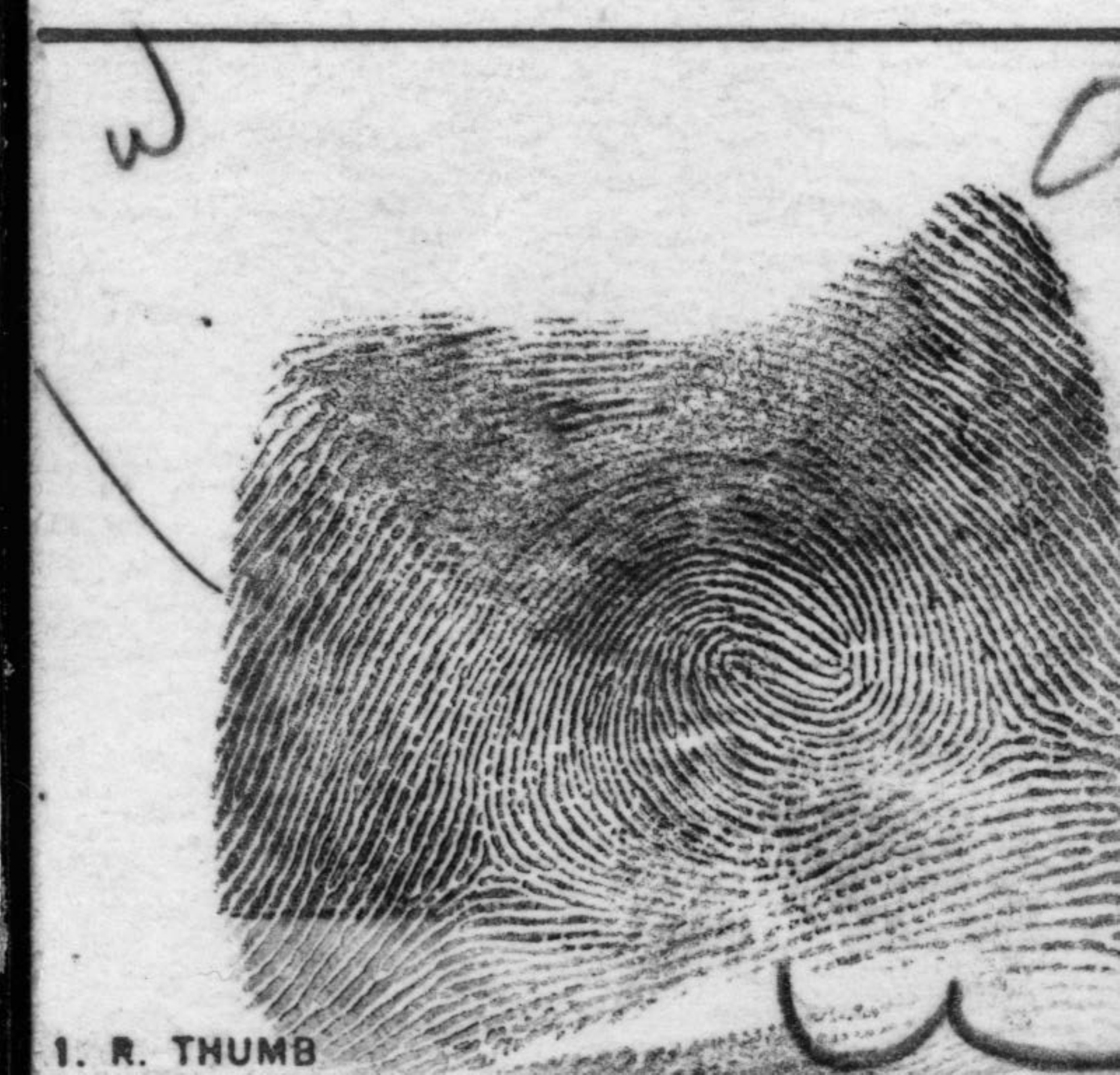}
}
\end{center}
   \caption{Example latents (first row) from NIST SD27 whose true rolled mates (second row) could not be retrieved at rank-1 by a leading COTS latent AFIS. This can be attributed to large background noise and poor quality ridge structure in (a), and small friction ridge area in (b). }
\label{fig:LatentEg}
\end{figure}

Over the past few years, deep networks, in particular, convolutional neural networks (ConvNets) have become the
dominant approach for addressing problems involving noisy, occluded and partial patterns and large numbers of classes. 
This is supported by state-of-the-art
performance of deep networks in large-scale image recognition \cite{ResNet}, unconstrained face recognition \cite{Taigman} and speech recognition in cluttered background \cite{HintonSpeech}, where traditional representation and  matching approaches fail. So, it is natural to consider ConvNets for latent fingerprint recognition.
 However, only a few published studies have applied 
 ConvNets to latent fingerprint recognition, and even these studies are limited to  individual modules, such as ridge flow estimation \cite{CaoICB2015} and minutiae extraction \cite{SankaranIJCB2014} \cite{Tang2016} of latent AFIS.  To our knowledge, there is no published study on designing a complete latent AFIS based on  ConvNets.

\subsection{Contributions}

In this paper, we design and build an automated latent recognition system and evaluate its performance against a leading latent AFIS. Meagher and Dvornychenko \cite{Lightsout} define seven tiers of possible latent print ``lights out" scenarios. They go on to say that ``for technical reasons,  only Tiers 1 and 2 are implementable now or in the near term.  Tiers 3 through 7 reflect our concept of an incremental approach to full lights-out capability." Our automated latent recognition system follows under Tier 2 of their definition where latent print experts submit latent searches and then receive the AFIS candidate list. All preprocessing, except region of interest (ROI), minutiae extraction, template generation and search has been automated. See Fig. \ref{fig:LatentROIEg}. 

The main contributions of this paper are as follows: 

\begin{enumerate}
 
	\item Input latent is represented by three different templates, each providing complementary  information. Two of the templates are minutiae-based, whereas the third template is texture-based. The minutiae in the minutiae-based templates are extracted, respectively, based on (i) ridge flow learned from a ConvNet, and (ii) dictionary of ridge structure elements. 

	\item Multi-scale and multi-location windows in the neighborhood of minutiae are used to learn minutiae descriptors. To develop salient minutiae descriptors, we train 14 different ConvNets, where each descriptor ConvNet is trained on a specific patch size at a specific location around the minutiae. A systematic feature selection (sequential forward selection) showed that only 3 out of the 14 ConvNets are adequate to maintain rank-1 recognition accuracy at significant computational savings. 
	
	\item Second order (minutiae pairs) and third order (minutiae triplets) graph-based minutiae correspondence algorithms are developed to minimize false minutiae correspondences in latent to its non-mate comparisons.
	
	\item A prototype of our latent recognition algorithm was evaluated on two different benchmark databases: NIST SD27 (258 latents) \cite{NISTDB27} and WVU latent DB (449 latents) \cite{WVU} against a reference database of 100,000 rolled prints.
	 The rank-1 retrieval for these two databases are:  64.7\%  for NIST SD27 and 75.3\% for WVU latent DB.  These results with automated preprocessing, feature extraction, and comparison are superior to published
	 results on these two databases. 
	 
	\item Score-level (rank-level) fusion of our algorithm with a leading  COTS latent AFIS, improves the  rank-1 accuracies to 73.3\% (74.4\%) for NIST SD27 and to 76.6\% (78.4\%) for WVU latent DB. This demonstrates that our approach to automated latent recognition based on ConvNets is complementary  to that used in the COTS latent AFIS.
\end{enumerate}

\section{Related Literature}

Given  a latent image, the main modules of a latent AFIS include  preprocessing (ROI segmentation, ridge flow estimation and ridge enhancement), feature (minutiae and texture) extraction and comparison. Fig. \ref{fig:LatentEg} shows challenges in latent processing: background noise, low contrast of friction ridge structure, and small friction ridge area. In the following, we briefly review major published algorithms pertaining to different modules. For a detailed review, see  \cite{Sankaran}. 

\begin{figure}[b]
	
	\begin{center}
		\captionsetup[subfigure]{labelformat=empty}
		\subfigure{
			\includegraphics[width=0.45\linewidth]{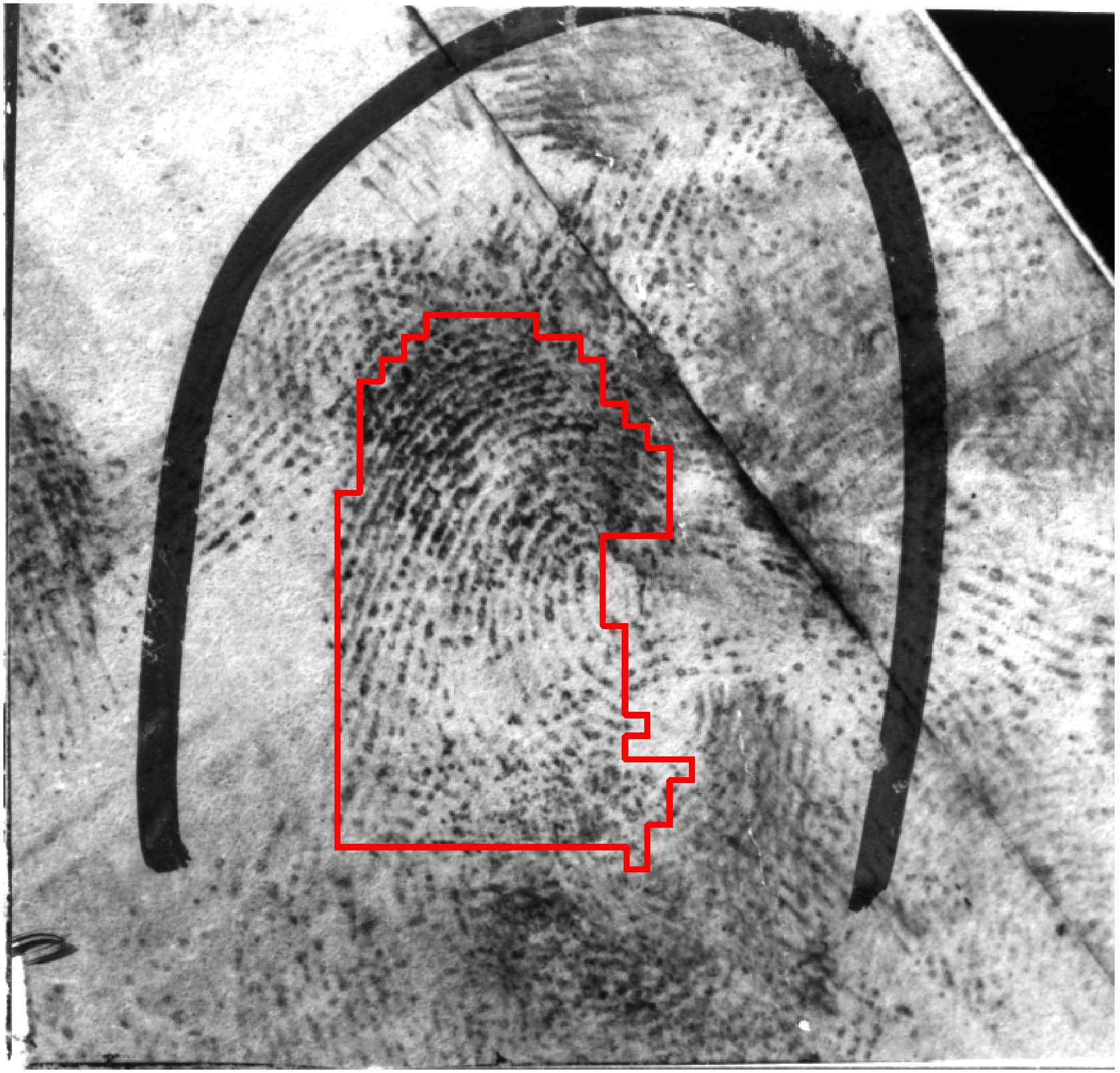}
		} \hspace{0.1cm}
		\subfigure{
			\includegraphics[width=0.45\linewidth]{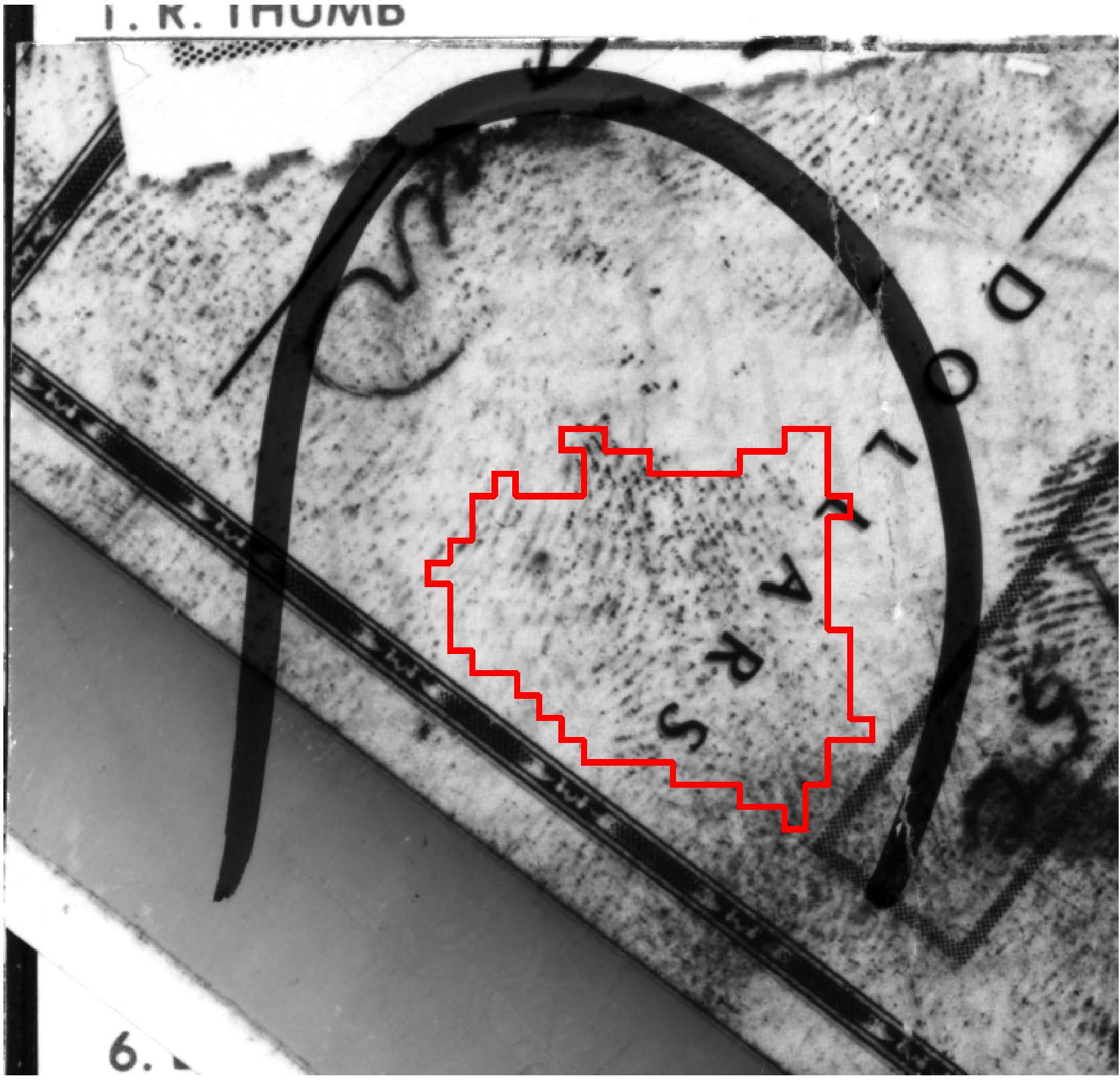}
		}
	\end{center}
	\caption{Latent fingerprints at a crime scene often contain multiple latent impressions, either of different individuals or multiple fingers of the same person. For this reason, a region of interest (ROI),  also called cropping,  outlined in red, is typically marked
		by examiners to highlight the friction ridge region of interest. }
	\label{fig:LatentROIEg}
\end{figure}

(a) \emph{ROI segmentation}. Published algorithms \cite{Karimi2008ICIP}, \cite{Short2011}, \cite{Zhang2013TIFS}, \cite{Choi2012}, \cite{CaoPAMI2014} do not work well on poor quality latents. Further, it is a common practice in forensics for an examiner to mark the ROI, also known as cropping (see Fig \ref{fig:LatentROIEg}), especially when there are overlapping latent impressions. We assume that ROI  for the query latent has been marked.

(b) \emph{Ridge flow estimation}. Two approaches for computing ridge flow have shown promise: (i) dictionary based learning, \cite{FengPAMI2014}, \cite{CaoPAMI2014}  and (ii)  ConvNet based learning \cite{CaoICB2015}. 
The ridge flow estimates from ConvNet generally perform better than dictionary based methods when evaluated against manually marked ridge flow \cite{CaoICB2015}. 

(c) \emph{Latent enhancement}. Gabor filtering is the most popular and effective approach, \cite{FengPAMI2014}, \cite{CaoPAMI2014}, \cite{CaoICB2015}. Other published approaches include multi-scale ridge dictionary using a set of Gabor elementary functions \cite{LiuTIFS2015}, and  a ridge dictionary with variable ridge and valley spacings \cite{CaoICB2016}.

 (d) \emph{Feature extraction}. A latent minutiae extractor using stacked denoising sparse autoencoder was proposed in \cite{SankaranIJCB2014}, but it showed poor performance on NIST SD27. While Cao \textit{et al.}  \cite{CaoICB2016} extracted minutiae, ridge clarity, singular point, and ridge orientation for automated latent value assessment, they did not integrate it with a latent matcher. Tang \textit{et al.} \cite{Tang2016}  developed a fully convolutional network for minutiae extraction, but it performed poorly compared to manually marked minutiae. 

 (e) \emph{Latent comparison}.  In the absence of a robust latent minutiae extractor, published latent comparison algorithms \cite{Feng2011PAMI},  \cite{PaulinoTIFS2013}, \cite{Krish2014ICPR}, \cite{Feng2017}   rely on manually marked minutiae.  

\begin{figure*}
	\begin{center}
		\includegraphics[width=0.85\linewidth]{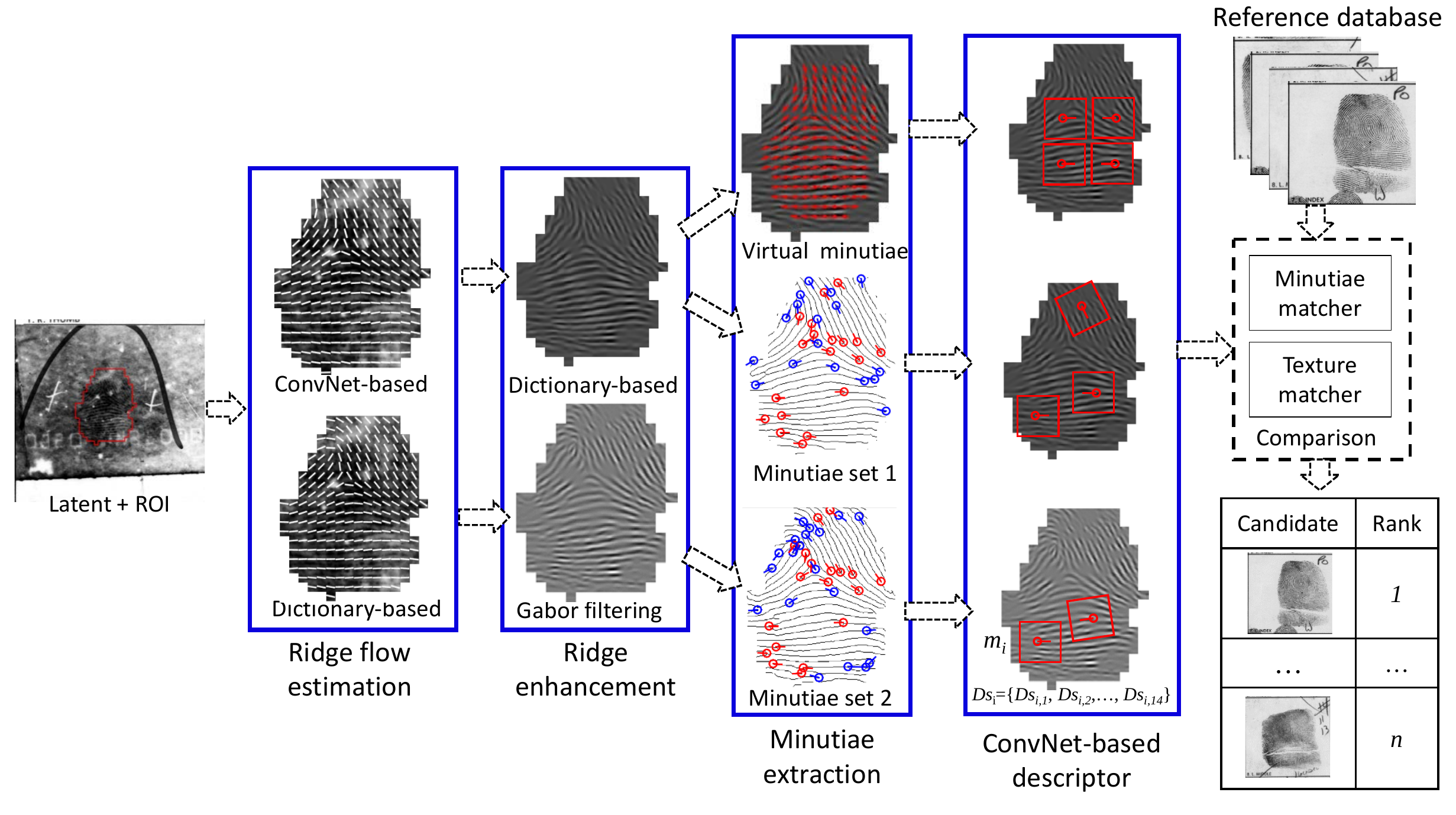}
	\end{center}
	\caption{Flowchart of the proposed latent recognition approach. The common minutiae in two true minutiae sets are shown in red. }
	\label{fig:overview}
\end{figure*}

In summary, to our knowledge, no automated latent recognition algorithm has been published in the literature. While ConvNets have been used for individual modules of a latent AFIS, their performance has not been evaluated in an end-to-end system. Even the number of available COTS latent AFIS is limited. In the 2012 NIST ELFT-EFS \#2 evaluation, 
there were only six  participants; the top three performers had significantly superior performance compared to the other three.  
The flowchart of the proposed latent recognition framework is shown in Fig. \ref{fig:overview}.  


\begin{figure}
	\begin{center}
		\includegraphics[width=0.95\linewidth]{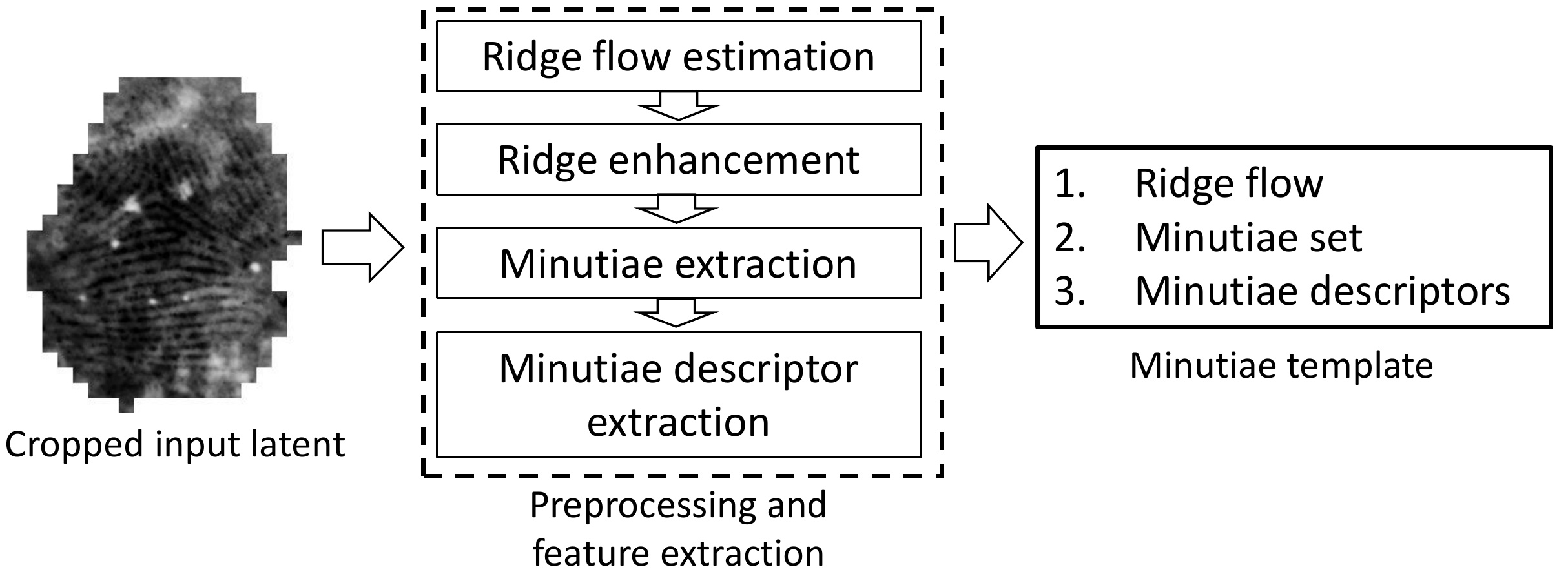}
	\end{center}
	\caption{Minutiae template generation. The same procedure is used for both minutiae template 1 and minutiae template 2.
	  }
	\label{fig:MinutiaeTemplate}
\end{figure}





\section{Preprocessing and Feature Extraction}
Latent feature extraction is presented in section \ref{sec:latent_feature}, where latent preprocessing is embedded into minutiae set extraction, and reference print feature extraction is provided in section 
\ref{sec:rolled_feature}.

\subsection{Latent Feature Extraction}
\label{sec:latent_feature}
Minutiae are arguably the most important features in fingerprint recognition. Two minutiae templates and one texture template are extracted for each latent (see Fig. \ref{fig:overview}).
While the two minutiae templates use the same framework (Fig. \ref{fig:MinutiaeTemplate}), they are based on different ridge flow estimation methods (ConvNet-based and Dictionary-based) and ridge enhancement methods (Dictionary-based and Gabor filtering-based). A minutiae template consists of ridge flow, a minutiae set (minutiae locations and orientations), and minutiae descriptors extracted by ConvNets using local latent patches. 

\subsubsection{Minutiae Set 1 }
\label{sec:minutiae_set_1}



The first minutiae set is extracted from the approach in \cite{CaoICB2016}, which consists of  the following steps: 1) ridge flow estimation using ConvNet, 2) ridge and valley contrast enhancement, 3) ridge enhancement by a ridge structure dictionary with variable ridge and valley spacing,  4) ridge binarization and thinning, and 5) minutiae detection in the skeleton image.

\subsubsection{Minutiae Set 2}

A coarse to fine dictionary is adopted to estimate ridge flow and ridge spacing \cite{CaoPAMI2014}. Gabor filtering tuned using the estimated ridge flow and ridge spacing is used to enhance the ridge structure.  Minutiae are then extracted from the enhanced latent to obtain  minutiae set 2.  A comparison in Fig. \ref{fig:overview} shows the complementary nature of minutiae sets 1 and 2.


\subsubsection{Texture Template}
\label{sec:latent_texture}
A texture template is introduced to account for situations where  the latent is of such a small area that it does not contain sufficient number of minutiae (for reliable comparison to reference prints) or the latent is of very poor quality so the minutiae extraction is not reliable.
In a texture template, we represent each non-overlapping local block ($s_b \times s_b$  pixels) in the latent by a pair of \textit{virtual minutiae}.
Let $(x,y)$ and $\alpha$ be the location and orientation of the center of a block. 
Then the virtual minutiae pair is located at $(x,y,\alpha)$ and $(x,y,\alpha+\pi)$. Note that the virtual minutiae do not correspond to ridge endings and bifurcations and the virtual minutiae close to the border are removed. The same minutia descriptor algorithm (section \ref{sec:descriptor}) used for the true minutiae sets is also used for virtual minutiae. The block size is set to $16\times 16$ to balance template efficacy and computational efficiency.

\subsubsection{Minutiae Descriptor }
\label{sec:descriptor}

\begin{figure*}
	\begin{center}
		\includegraphics[width=0.7\linewidth]{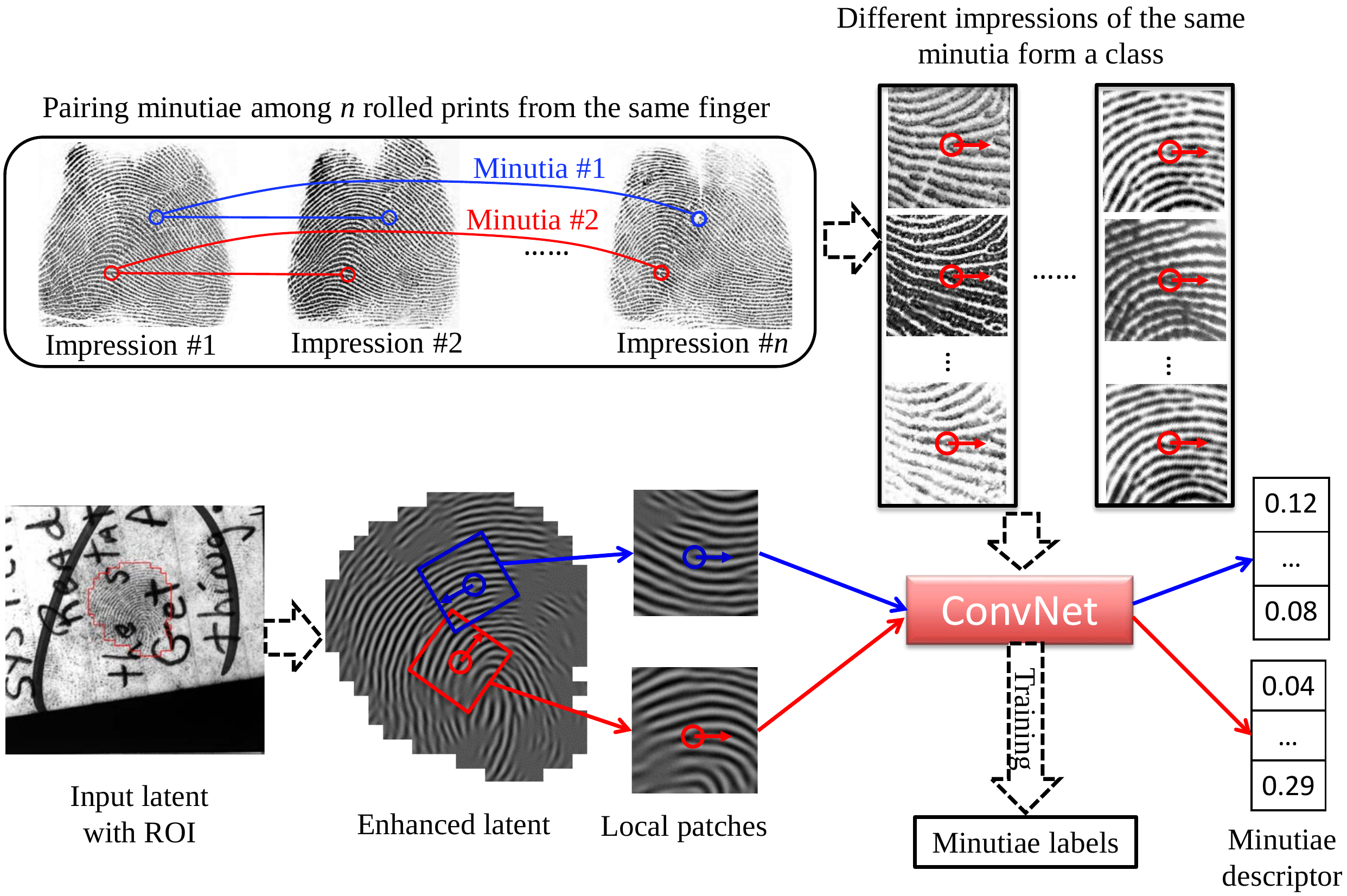}
	\end{center}
	\caption{Minutiae descriptor extraction via ConvNet. The dotted arrows show the offline training
		process, while solid arrows show the online process for minutiae descriptor extraction. A total of 800K fingerprint patches from 50K minutiae, extracted from the MSP longitudinal fingerprint database \cite{Yoon2015PNAS}, were used for training the ConvNet. The patch size shown here is $80\times80$ pixels.}
	\label{fig:DescriptorFlowchart}
\end{figure*}

\begin{figure}[t]
	
	\begin{center}
		\subfigure[]{
			\includegraphics[scale=0.3]{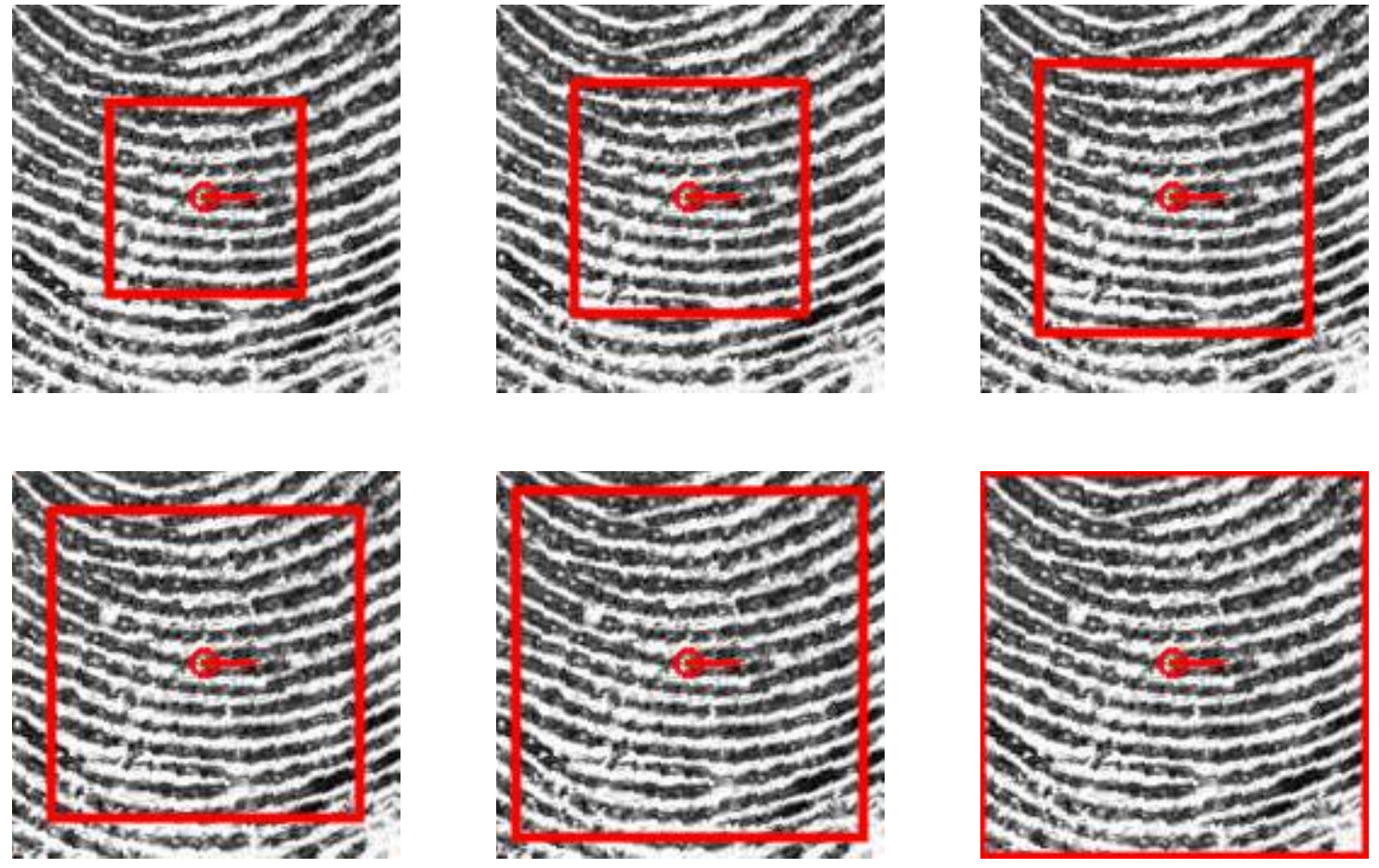}
		}\hspace{0.3cm}
		\subfigure[]{
			\includegraphics[scale=0.3]{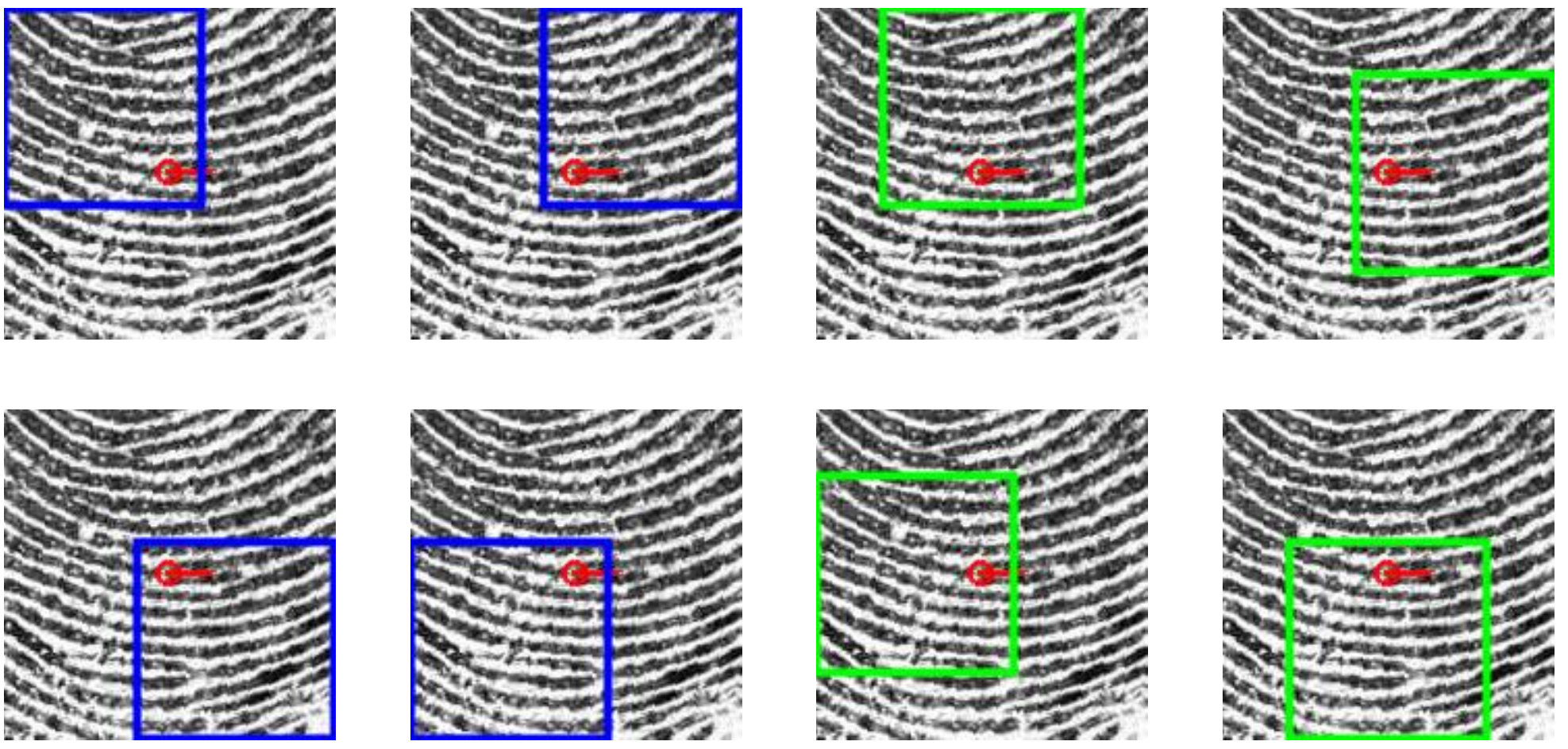}
		}\hspace{0.1cm}
	\end{center}
	\caption{Fourteen types of fingerprint patches, with different size and location, centered at a minutia (shown in red). 
		 Patches at (a) 6 different scales and (b) in 8 different locations  around minutia:  top left, top right, bottom right and bottom left, top, right, left and bottom. The fingerprint patches shown here are of size $160\times 160$ pixels. The window sizes (scale) in (a) are $80\times 80$, $96\times 96$, $112\times 112$, $128\times 128$, $144\times 144$, and $160\times 160$ pixels. The windows in (b) are all of size $96\times 96$ pixels. }
	\label{fig:PatchTypes}
\end{figure}

A minutia descriptor contains attributes of the minutia based on the image characteristics in its neighborhood. Salient descriptors are needed to eliminate false minutiae correspondences between a latent and reference prints. Instead of specifying the descriptor in an ad hoc manner \cite{Feng2011PAMI}, we  train ConvNets to learn minutiae descriptor from local fingerprint patches.  As demonstrated in face recognition, for example,  \cite{Sun2014CVPR}, training a set of ConvNets using multiple image patches at different scales and regions can significantly boost the recognition performance. In this paper,  we adopt a multi-scale approach, where fingerprint patches of different sizes and at different locations (a total of 14 patches)  are defined as shown in Fig. \ref{fig:PatchTypes}. Multiple instances of patches extracted for the same minutia are used to train 14 different ConvNets\footnote{The toolbox MatConvNet \cite{MatConvNet} is used to implement the ConvNet architecture. Offline training of the ConvNet is conducted on a Linux server with Tesla K20 GPUs.}.   
  The flowchart of minutiae descriptor extraction for one of the 14 ConvNets is illustrated  in Fig. \ref{fig:DescriptorFlowchart}. 
 The details are as follows.
 \begin{enumerate}
\item \emph{Training patch selection}. Multiple patches around the same minutiae extracted from different fingerprint impressions of the same finger  are needed. For this purpose, we utilize MSP longitudinal fingerprint database\footnote{No longitudinal latent database is available for  training descriptor ConvNet.} \cite{Yoon2015PNAS}, which 
contains  1,311 subjects with at least 10 rolled impressions, collected over at least 5 years, with a total of 165,880 fingerprints.  Only those minutiae in these prints which can be extracted in eight or more impressions of the same finger are retained for training. This ensures that we are only using reliable minutiae. Local fingerprint patches around these selected minutiae are extracted to train the ConvNets. 

\item \emph{Training}. 
 We adopt the same ConvNet architecture in \cite{CaoICB2015} for all 14 patch types. Smaller patches are resized to $160\times 160$ pixels using bilinear interpolation to ensure that we can use  the same ConvNet \cite{CaoICB2015} with  $160\times 160$  images as input.  
  Random shifts (-5 to 5 pixels) and rotations (-5$^{\circ}$  to 5$^{\circ}$) of the patches are used to augment the training set.

\item \emph{Latent minutiae descriptor extraction}. For each ConvNet, its 128-dimensional output of the last fully connected layer is considered as a feature vector. A minutia descriptor could be a concatenation of a subset of the 14 feature vectors output by the 14 ConvNets. 
\end{enumerate}

\subsection{Reference print Feature Extraction}
\label{sec:rolled_feature}
Reference prints are typically of higher quality compared to latents, so it is easier to get reliable minutiae from them. For this reason, we extract only one minutiae template, but we still extract the texture template. 
The reference print minutiae are extracted by a COTS tenprint AFIS rather than the proposed minutiae extractor for latents. The ridge flow is extracted by Short Time Fourier Transform (STFT) \cite{Chikkerur2007198}. A reference print minutiae template, similar to latents, includes (i) ridge flow, (ii) minutiae set and (iii) minutiae descriptors (section \ref{sec:descriptor}).

The texture template for reference print is extracted in a manner similar to latents (section \ref{sec:latent_texture}).
For computational efficiency, each nonoverlapping  block of $s_b\times s_b$ pixels is considered to define a single virtual minutia. On average, there are 1,018 virtual minutiae in a reference print. The texture template consists of  a virtual minutiae set, and their descriptors (section \ref{sec:descriptor}).   
 Since the latent texture template considers two virtual minutiae, we expect that at least one of them will be in correspondence with the reference print virtual minutia in the true mate.

\section{Latent to Rolled Comparison}
Two latent-to-reference print comparison algorithms are designed: (i) a minutiae template comparison algorithm and (ii) a texture template comparison algorithm.

\subsection{Minutiae Template Comparison}

\label{sec:MinutiaeMatching}


Let $M^l = \{m_i^l =(x_i^l,y_i^l,\alpha_i^l )\}_{i=1}^{n_l} $ denote the latent minutiae set with $n_l$ minutiae, where $(x_i^l,y_i^l)$  and 
$\alpha_i^l$ are the location and orientation of the $i^{th}$ minutia, respectively. Let  $M^r = \{m_j^r =(x_j^r, y_j^r,\alpha_j^r)\}_{j=1}^{n_r}$  denote a reference print minutiae set with $n_r$ minutiae,  where $(x_j^r,y_j^r)$ and $\alpha_j^r$ are the location and orientation  of the $j^{th}$ rolled minutia, respectively.  
The  minutiae template comparison algorithm seeks to establish the minutiae correspondences between $M^l$ and $M^r$.  We impose the constraint that no minutia in one set should match more than one minutia in the other set. The problem of minutiae correspondence can be formulated as an optimization problem to find the assignment $X \in \mathbb{S}$, where:
\begin{equation}
\small
\mathbb{S} = \{ X\in \{0,1\}^{n_l\times n_r}, \forall i, \sum\limits_{i}X_{i,j}\leqslant 1 \nonumber ,  \forall j,  \sum\limits_{j}X_{i,j}\leqslant 1 \}, 
\end{equation} 
$X_{i,j} = 1$ if $m_{i}^l$ and $m_{j}^r$  are in correspondence and $X_{i,j} = 0$, otherwise.

In the second-order graph based minutiae correspondence algorithm \cite{Fu2013ICB},  the objective function $S_2$ is defined as:
\begin{equation}
\label{eq:objective_two}
S_2(X) = \sum_{i_1,i_2,j_1,j_2}  \\ H^2_{i_1,i_2,j_1,j_2} X_{i_1,i_2}X_{j_1,j_2},
\end{equation}
where $H^2\in \mathbb{R}^{n_l\times n_r\times n_l\times n_r}$ is a 4-dimensional tensor and  $H^2_{i_1,i_2,j_1,j_2}$ measures the compatibility between latent minutiae pair $(m^l_{i_1},m^l_{j_1})$  and rolled  minutiae pair $(m^r_{i_2},m^r_{j_2})$. 

One limitation of the second-order graph matching (or pairwise minutiae correspondence) is that it is possible that two different minutiae configurations may have similar minutiae pairs. To circumvent this, higher order graph matching, has been proposed  to reduce the number of false correspondences \cite{Duchenne2011}. Here, we consider the third-order graph matching (minutiae triplets) whose objective function is given as:
\begin{equation}
\label{eq:objective_three}
\small
  S_3(X) = \sum_{i_1,j_1,k_1,i_2,j_2,k_2} \\ H^3_{i_1,i_2,j_1,j_2,k_1,k_2} X_{i_1,i_2}X_{j_1,j_2}X_{k_1,k_2},
\end{equation}
where $H^3\in \mathbb{R}^{n_l\times n_r\times n_l\times n_r\times n_l\times n_r}$ is a 6-dimensional tensor  and $H^3_{i_1,i_2,j_1,j_2,k_1,k_2}$ measures the compatibility between latent minutiae triplet $(m^l_{i_1},m^l_{j_1},m^l_{k_1})$ and reference print minutiae triplet $ (m^r_{i_2},m^r_{j_2},m^r_{k_2})$.  Since $H^3$ is of size $(n_l\cdot n_r)^3$ and $H^2$ is of size $(n_l\cdot n_r)^2$, this approach is more computationally demanding than the second-order graph matching.

\subsubsection{ Proposed Minutiae Correspondence Algorithm}
Minutiae descriptors allow us to consider only a small subset of minutiae correspondences among the $n_l\times n_r$ possible correspondences. For computational efficiency, only the top $N$ ($N=120$) minutiae correspondences  are selected based on their descriptor similarities. Since the second-order graph matching is able to remove most of the false correspondences, we first use the second-order graph matching, followed by the third-order graph matching for minutiae correspondence.  \textbf{Algorithm} \ref{alg:pairing} shows the main steps of the proposed minutiae correspondence algorithm.

\begin{algorithm}
	\caption{Minutiae correspondence algorithm}\label{alg:pairing}
	\begin{algorithmic}[1]
		\State \textbf{Input:} Latent minutiae template with $n_l$ minutiae and reference print minutiae template with $n_r$ minutiae
		\State \textbf{Output:} Minutiae correspondences
		\State 	Compute the $n_l\times n_r$ minutiae similarity matrix using Eq. (\ref{eq:descriptor_similarity})\;
		\State	Select the top $N$ minutiae correspondences based on the above minutiae similarity matrix\;
		\State 	Construct $H^2$ based on these $N$ minutiae pairs\;
		\State	Remove false minutiae correspondences using \textbf{Algorithms} \ref{alg:secondorder} and \ref{alg:Discretization}\;
		\State	Construct $H^3$ for the remaining minutiae pairs\;
		\State	Remove false minutiae correspondences using \textbf{Algorithms} \ref{alg:thirdorder}  and \ref{alg:Discretization}\;
		\State 	Output final minutiae correspondences.
	\end{algorithmic}
\end{algorithm}

In the following, we first present how to construct  $H^2$ and $H^3$ and  then give details of the minutiae correspondence algorithm.

\subsubsection{Construction of $H^2$ and $H^3$}
\begin{figure}[h]
	\begin{center}
		\subfigure[]{
			\includegraphics[width=0.35\linewidth]{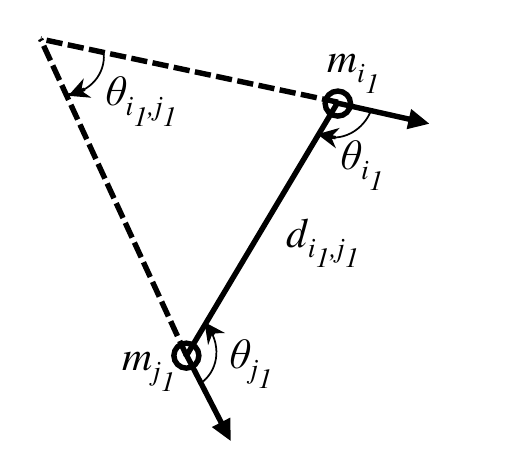}
		}\hspace{0.1cm}
		\subfigure[]{
			\includegraphics[width=0.45\linewidth]{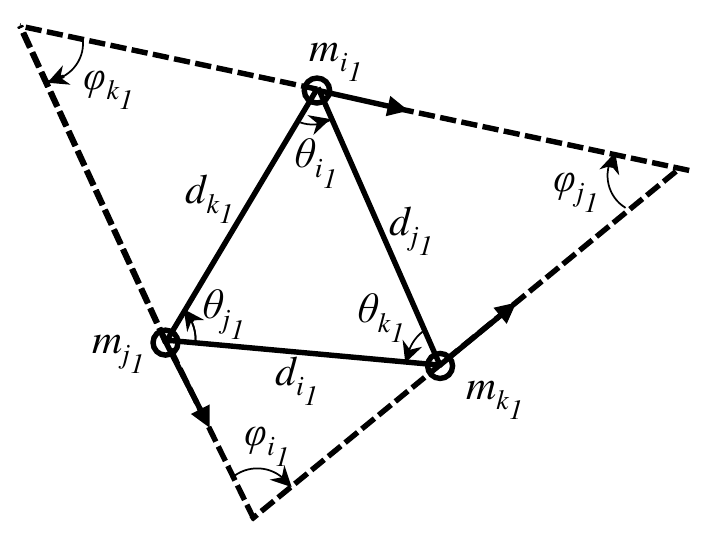}
		}\hspace{0.1cm}
	\end{center}
	\caption{Illustration of feature representation of (a) a minutiae pair  $(m_{i_1} , m_{j_1})$ and (b) a minutiae triplet $(m_{i_1} , m_{j_1},  m_{k_1})$,  where the solid arrows denote minutiae orientations.  }
	\label{fig:representation}
\end{figure}

The term $H^2_{i_1,i_2,j_1,j_2}$ in Eq. (\ref{eq:objective_two})  measures the compatibility between a minutiae pair $(m^l_{i_1},m^l_{j_1})$ of the latent and a minutiae pair  $(m^r_{i_2},m^r_{j_2})$ of the
reference print. 
A 4-dimensional feature vector is computed to characterize each minutiae pair. Let ($d_{i_1,j_1}$, $\theta_{i_1}$, $\theta_{j_1}$, $\theta_{i_1,j_1}$) and ($d_{i_2,j_2}$, $\theta_{i_2}$, $\theta_{j_2}$, $\theta_{i_2,j_2}$) denote two feature vectors for a minutiae pair from a  latent and a reference print, respectively. 
Fig. \ref{fig:representation} (a) illustrates the feature vector. $H^2_{i_1,i_2,j_1,j_2}$ 
is computed as:
\begin{align}
H^2_{i_1,i_2,j_1,j_2} &= \Pi_{p=1}^{4}Z(d_p,\mu_p,\tau_p,t_p),
\end{align}
where
\begin{align}
d_1 &= |d_{i_1,j_1} - d_{i_2,j_2}  |,  \label{eq:eculid} \\
d_2 &= \min( |\theta_{i_1} - \theta_{i_2} |,   2\pi-|\theta_{i_1} - \theta_{i_2} |), \label{eq:directional_1}  \\
d_3 &= \min( |\theta_{j_1} - \theta_{j_2} |,   2\pi-|\theta_{j_1} - \theta_{j_2} |), \label{eq:directional_2}  \\
d_4 &= \min( |\theta_{i_1,j_1} - \theta_{i_2,j_2} |,   2\pi-|\theta_{i_1,j_1} - \theta_{i_2,j_2} |),  \label{eq:directional_3}
\end{align}
$Z$ is a truncated sigmoid function, which is defined as:
\begin{equation}
Z(v,\mu_p,\tau_p, t_p) =	\left.
\begin{cases}
 \frac{1}{1+e^{-\tau_p(v-\mu_p)}}, & \text{   if } v \leq t_p, \\
 0, & \text{   otherwise. }
\end{cases}\right.
\end{equation}
and $\mu_p,\tau_p$ and $t_p$ are parameters of function $Z$.


\begin{algorithm}
	\caption{Power iteration for the second-order eigenvalue problem}\label{alg:secondorder}
	\begin{algorithmic}[1]
		\State \textbf{Input:} Matrix $H^2$
		\State \textbf{Output:} $Y$, principal eigenvector of $H^2$
		\State 	Initialize $Y$ with small random positive numbers\;
		\While  {no convergence}{
		\State		$Y$ $\gets$ $HY$\;
		\State		$Y$ $\gets$ $\frac{1}{||Y||_2}Y$\;
		\EndWhile }
	\end{algorithmic}
\end{algorithm}

\begin{algorithm}
	\caption{Power iteration for the third-order eigenvalue problem}\label{alg:thirdorder}
	\begin{algorithmic}[1]
		\State \textbf{Input:} Matrix $H^2$
		\State \textbf{Output:} $Y$, principal eigenvector of $H^2$
		\State 	Initialize $Y$ with small random positive numbers\;
		\While  {no convergence}{
		\For{$i$}
		\State $Y_i$ $\gets$ $\sum_{j,k}H^3_{i,j,k}Y_jY_k$\;
		\EndFor 
		\State			$Y$ $\gets$ $\frac{1}{||Y||_2} Y$\;
		\EndWhile }
	\end{algorithmic}
\end{algorithm}

\begin{algorithm}
	\caption{Discretization to ensure a one-to-one matching}\label{alg:Discretization}
	\begin{algorithmic}[1]
		\State \textbf{Input:} Eigenvector Y output by {\textbf{Algorithms}} \ref{alg:thirdorder} or \ref{alg:secondorder}
		\State \textbf{Output:} Minutiae correspondences $C$
		\State 	Initialize threshold $T$ \;
		\State 	Initialize minutiae pair $C=\{\}$  \;
		\State	Set $flag_l(p)=0$, $p=1,2,...,n_l$ \;
		\State  set $flag_r(q)=0$, $q=1,2,...,n_r$ \;
		\While {$\max(Y) >T$}
			\State	$i =\arg\max(Y)$\;
			\State  Y(i) = 0\;
			\If{$flag_l(i_1)==1$ or $flag_r(i_2)==1$}
			\State \textbf{continue}\;
			\Else 	
			\State 	C.append($i_1$,$i_2$)\;
			\State 	$flag_l(i_1)=1$\;
			\State 	$flag_r(i_2)=1$\;
			\EndIf
		\EndWhile 		
	\end{algorithmic}
\end{algorithm}

\begin{figure}[t]
	\begin{center}
		\subfigure[]{
			\includegraphics[width=0.85\linewidth]{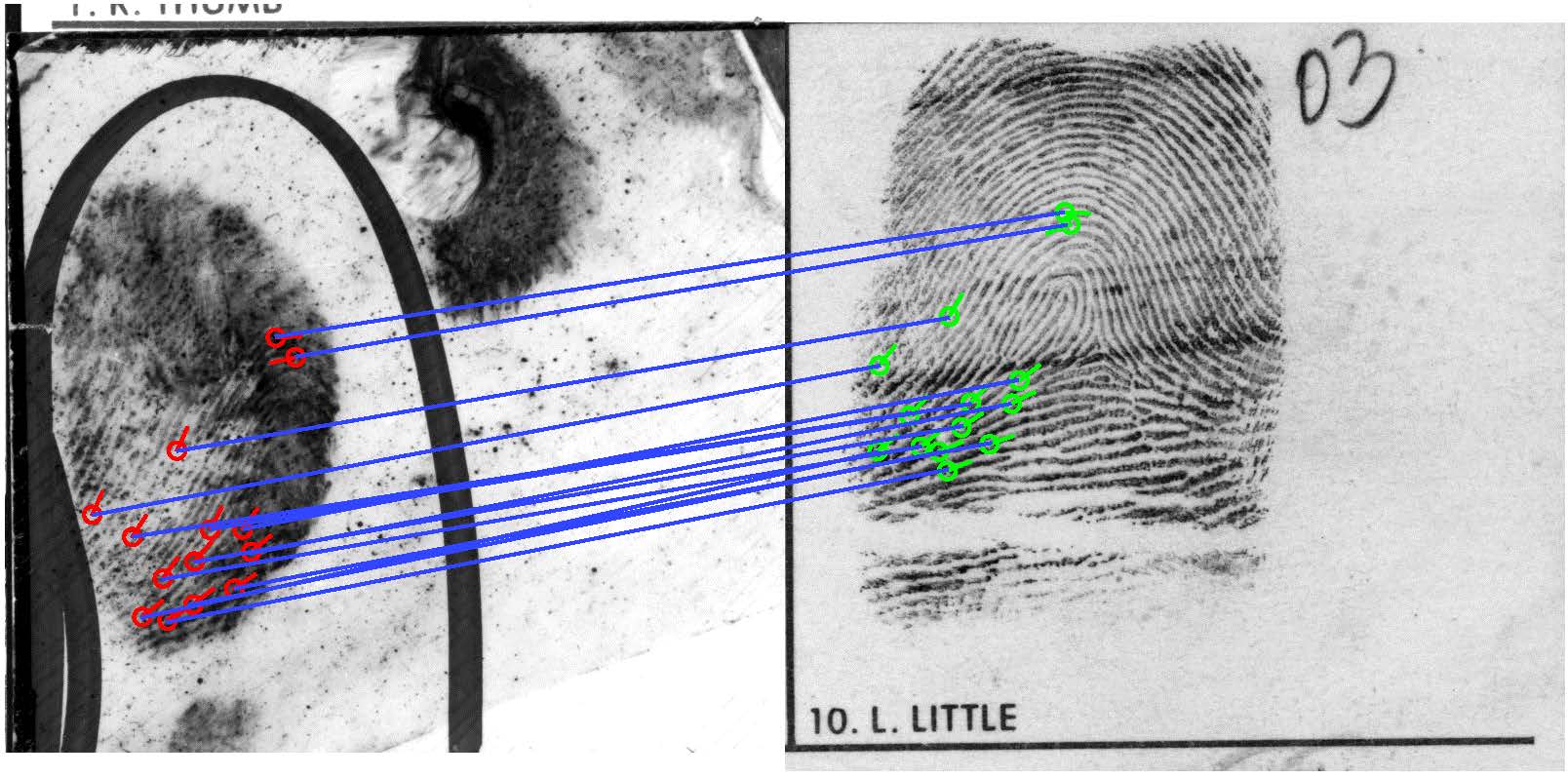}
		}\hspace{0.1cm}
		\subfigure[]{
			\includegraphics[width=0.85\linewidth]{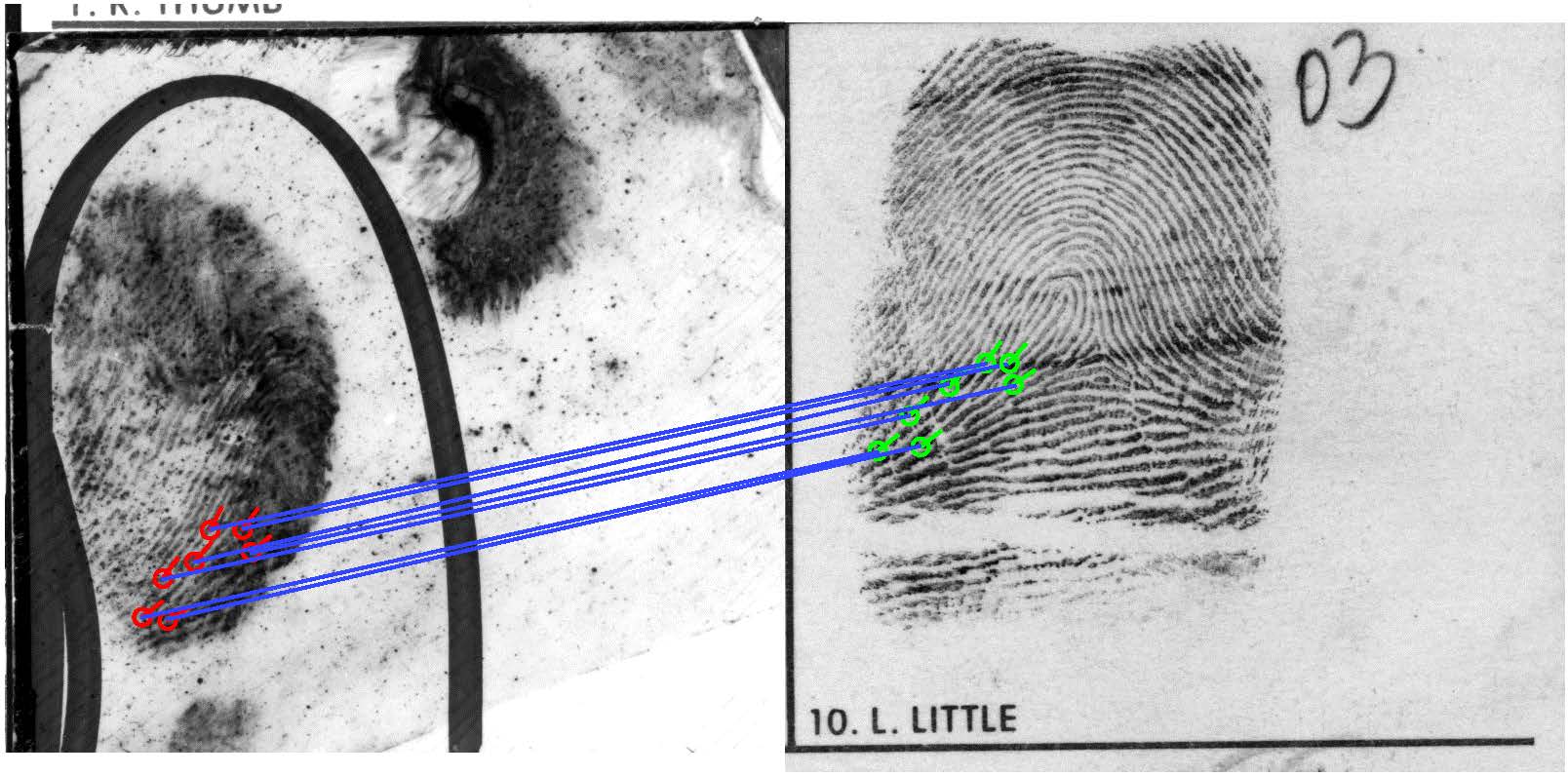}
		}\hspace{0.1cm}
		\subfigure[]{
			\includegraphics[width=0.85\linewidth]{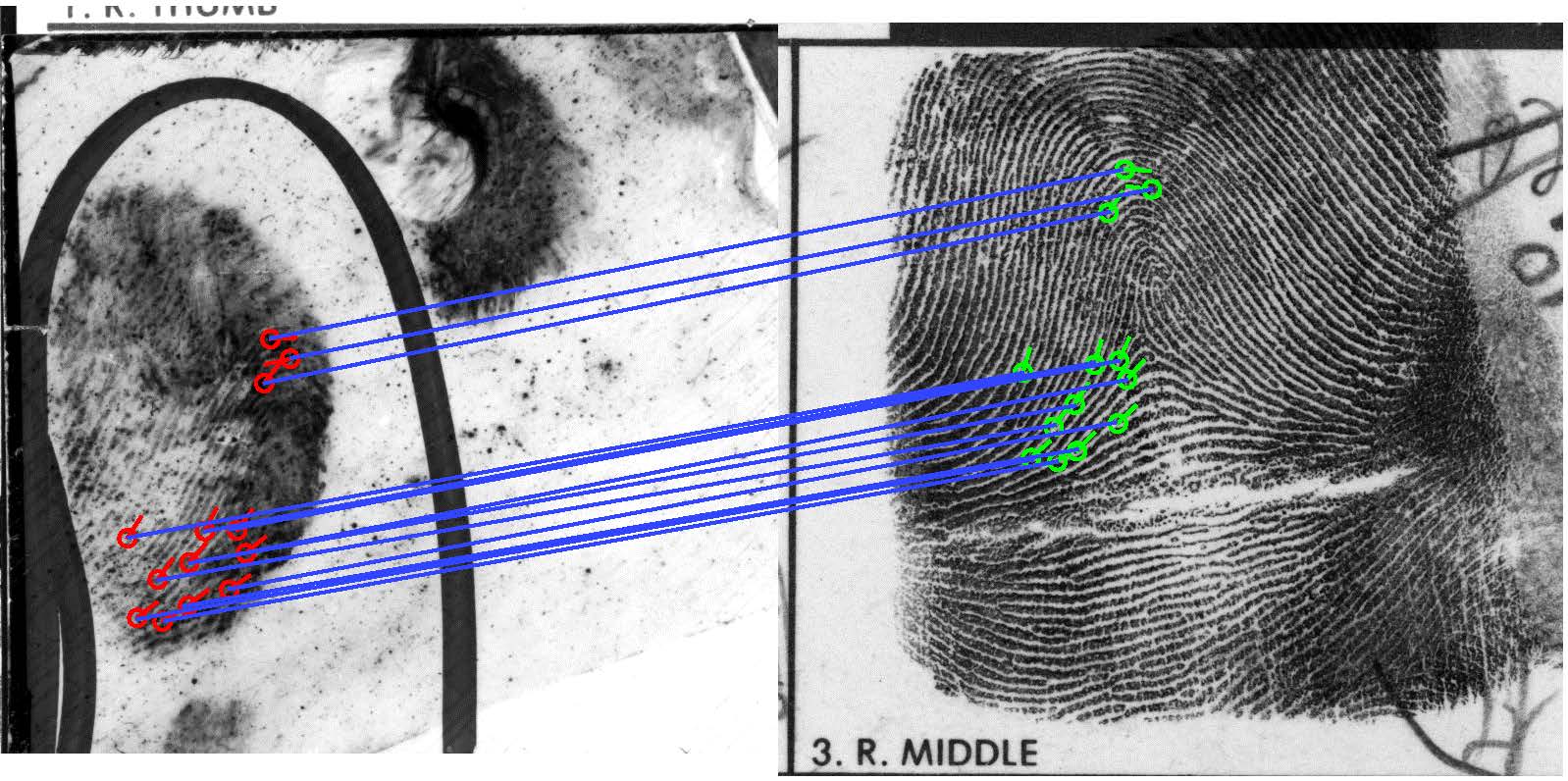}
		}\hspace{0.1cm}
	\end{center}
	\caption{Comparison of minutiae correspondences. (a) 14 minutiae pairs found in correspondence  between the latent and a non-mate \cite{PaulinoTIFS2013},  (b) 7 minutiae pairs found in correspondence for the same comparison as in (a) by the proposed method and (c) 13 minutiae pairs found in  correspondences between the latent and its true mate by the proposed method. Note that we use manually marked minutiae and MCC descriptor \cite{Cappelli2010PAMI} for a fair comparison with  \cite{PaulinoTIFS2013}. }
	\label{fig:MinutiaePairs}
\end{figure}

The term $H^3_{i_1,i_2,j_1,j_2,k_1,k_2} $ in Eq. (\ref{eq:objective_three}) measures the compatibility between a minutiae triplet $(m^l_{i_1},m^l_{j_1},m^l_{k_1})$ of the latent and a minutiae triplet  $(m^r_{i_2},m^r_{j_2},m^r_{k_2})$ of the
reference print. A 9-dimensional vector is computed to  characterize each minutiae triplet, as illustrated in Fig. \ref{fig:representation} (b). Let $(d_{i1},d_{j1},d_{k1},\theta_{i1},\theta_{j1},\theta_{k1}, \varphi_{i1},\varphi_{j1},\varphi_{k1}) $ and  $(d_{i2},d_{j2},d_{k2},\theta_{i2},\theta_{j2},\theta_{k2}, \varphi_{i2},\varphi_{j2},\varphi_{k2}) $ denote two feature vectors corresponding to the  two minutiae triplets from the latent and the reference print, respectively. 
 Then $H^3_{i_1,j_1,i_2,j_2,k_1,k_2}$  is computed as:
\begin{align}
\small
H^3_{i_1,j_1,i_2,j_2,k_1,k_2} &= \Pi_{p=i,j,k}\Pi_{q=1}^{3}Z(d_{pq},\mu_{pq},\tau_{pq},t_{pq}),
\end{align}
where
\begin{align}
d_{p1} &= |d_{p1} - d_{p2}  |,  \nonumber \\
d_{p2} &= \min( |\theta_{p1} - \theta_{p2} |,   2\pi-|\theta_{p1} - \theta_{p2} |), \nonumber \\
d_{p3} &= \min( |\phi_{p1} - \phi_{p2} |,   2\pi-|\phi_{p1} - \phi_{p2} |), \nonumber \\
p& = i,j,k. \nonumber
\end{align}
There are two kinds of distances used in computing $H^2$ and $H^3$, i.e. Euclidean distance (e.g., Eq. (\ref{eq:eculid})) between minutiae locations,  and directional distance (e.g., Eqs. (\ref{eq:directional_1}), (\ref{eq:directional_2}) and (\ref{eq:directional_3}) )   between minutiae angles.  For the Euclidean distance, $\mu, \tau$ and $t$  are set as 15, $-\frac{1}{5}$ and 40, respectively.  For the directional distance, $\mu, \tau$ and $t$ are set as $\frac{1}{12}$, $-15$ and $\pi/4$, respectively.  These tolerance values were determined empirically.

\subsubsection{Proposed Minutiae Correspondence} 
Suppose $Des_{i_1}^l = \{Des_{i_1}^l(p)\}_{p\in P} $ and  $Des_{i_2}^r = \{Des_{i_2}^r(p)\}_{p\in P} $ are two sets of minutia descriptors of the ${i_1}$th latent minutia  and the ${i_2}$th reference print minutia, respectively, where $P$ is a subset of the 14 ConvNets. The descriptor similarity $DesSim(i_1,i_2)$ between $Des_{i_1}^l$ and $Des_{i_2}^r$ is computed based on cosine distance as follows:
\begin{equation}
\label{eq:descriptor_similarity}
\small
DesSim(i_1,i_2) =\frac{1}{\sum_{p\in P} 1 } \sum_{p\in P} \frac{(Des_{i_1}^l(p))^T\cdot Des_{i_2}^r(p)}{||Des_{i_1}^l(p)||\cdot ||Des_{i_2}^r(p)||}.
\end{equation}
As in section 4.1.2, the top $N$ minutiae correspondences with the highest similarity values in (\ref{eq:descriptor_similarity}) are selected.   Suppose $\{(i_1,i_2)\}_{i=1}^{N}$ are the $N$ selected  minutiae pairs, and $Y$ is an $N$-dimensional correspondence vector, where the $i^{th}$ element ($Y_i$) indicates whether $i_1$  is assigned to $i_2$  ($Y_i = 1$) or not ($Y_i = 0$). The objective function in (\ref{eq:objective_two}) can be simplified as 
\begin{equation}
\label{eq:objective_simple_two}
S_2(Y) = \sum_{i,j}  \\ H^2_{i,j} Y_i Y_j,
\end{equation}
where  $i=(i_1,i_2)$ and  $j=(j_1,j_2)$ are  two selected minutiae correspondences, and $H^2_{i,j} $  is  equivalent to  $H^2_{i_1,j_1,i_2,j_2}$. 
Objective function (\ref{eq:objective_three})  can be similarly  rewritten as:
\begin{equation}
\label{eq:objective_simple_three}
S_3(Y) = \sum_{i,j,k}  \\ H^3_{i,j,k} Y_i Y_jY_k,
\end{equation}
where $i=(i_1,i_2)$,  $j=(j_1,j_2)$ and $k=(k_1,k_2)$  are three selected minutiae correspondences,  and $H^3_{i,j,k} $ is equivalent to $H^3_{i_1,i_2,j_1,j_2,k_1,k_2}$. 

The second-order graph matching problem (\ref{eq:objective_simple_two}) is a quadratic assignment problem, with no known polynomial time algorithm for solving it. This also holds for the third-order graph matching problem (\ref{eq:objective_simple_three}). 
A strategy of power iteration, followed by discretization \cite{Duchenne2011} is a simple but efficient approach to obtain approximate solution for (\ref{eq:objective_simple_two}) and (\ref{eq:objective_simple_three}). The power iteration methods for (\ref{eq:objective_simple_two}) and (\ref{eq:objective_simple_three}) are shown in \textbf{Algorithms} \ref{alg:secondorder} and \ref{alg:thirdorder}, respectively. \textbf{Algorithm} \ref{alg:Discretization} is the discretization step to ensure  a one-to-one matching.


Figs. \ref{fig:MinutiaePairs} (a) and (b) compare the proposed minutiae correspondence algorithm with the method of \cite{PaulinoTIFS2013} on an impostor comparison (latent to a non-mate comparison). Fig. \ref{fig:MinutiaePairs} (c) shows an example of minutiae correspondences for a genuine match  between a latent and its rolled mate. 

\subsubsection{Minutiae Template Similarity}
The similarity between a latent minutiae template and a reference minutiae template  consists of two parts: (i) minutiae similarity, i.e., similarity of descriptors of matched minutiae correspondences, and (ii) ridge flow similarity. Suppose  $\{(m^l_{i_1}=(x_{i_1}^l,y_{i_1}^l,\alpha_{i_1}^l),m^r_{i_2}=(x_{i_2}^r,y_{i_2}^r,\alpha_{i_2}^r))\}_{i=1}^{n}$ are the $n$ matched minutiae correspondences between the latent and the reference print by \textbf{Algorithm} \ref{alg:pairing}. The minutiae similarity $S_M$ is defined as:
\begin{equation}
\label{eq:minutiae_similarity}
S_M = \sum_{i=1}^nDesSim(i_1,i_2),
\end{equation}
where $DesSim(i_1,i_2)$ is the descriptor similarity between $Des_{i_1}^l$  and  $Des_{i_2}^r$ in Eq. (\ref{eq:descriptor_similarity}).
The ridge flow similarity is computed by first aligning the two ridge flow maps using the minutiae correspondences and then computing the orientation similarity of overlapping blocks. The rotation $\Delta \alpha$, and translation $(\Delta x, \Delta y)$ is computed as:
\begin{align}
\label{eq:alignment}
\Delta \alpha &= \arctan(\sum_{i=1}^n\sin(\Delta \alpha_i ),\sum_{i=1}^n\cos(\Delta \alpha_i)), \\
\Delta x &= \frac{1}{n}\sum_{i=1}^n ( x_{i_2}^r - x_{i_1}^l\cos(\Delta \alpha) + y_{i_1}^l\sin(\Delta \alpha)),\\
\Delta y &= \frac{1}{n}\sum_{i=1}^n ( y_{i_2}^r - y_{i_1}^l\cos(\Delta \alpha) - x_{i_1}^l\sin(\Delta \alpha)),
\end{align}
where $\Delta \alpha_i = (\alpha_{i_2}^r-\alpha_{i_1}^l)$. The values of  $\Delta \alpha$ and $(\Delta x, \Delta y)$ are  used for ridge flow alignment. Let $\{O_{k,1}\}_{k=1}^K$ and $\{O_{k,2}\}_{k=1}^K$ denote the orientations in the overlapping $K$ blocks for the latent and the reference print, respectively. The ridge flow similarity $S_O$ is given by 
\begin{equation}
\label{eq:OS}
S_O = \frac{1}{K}|\sum_{k=1}^K e^{(2\sqrt{-1} (O_{k,1} - O_{k,2}))}|.
\end{equation}
The minutiae template similarity $S_{MT}$ is computed as  the product of the minutiae similarity and ridge flow similarity,
\begin{equation}
\label{eq:MTS}
S_{MT}= S_M\cdot S_O.
\end{equation}

\subsection{Texture Template Similarity}
The same minutiae comparison algorithm proposed in section \ref{sec:MinutiaeMatching} can be used for virtual minutiae comparison  in texture template. However, there are two main differences: (i) top $N=200$  virtual minutiae correspondences, rather than 200 for real minutiae, are selected based on descriptor similarity, and (ii) the texture template similarity  $S_{TT}$ only consists of the sum of the similarities of matched virtual minutiae correspondences in Eq. (\ref{eq:minutiae_similarity}).


\subsection{Similarity Score Fusion}
Two minutiae templates and one texture template are extracted for each latent,  but only  one minutiae template and one texture template are extracted for each reference print. Two  minutiae template similarity scores  ($S_{MT,1}$ and $S_{MT,2}$)  are generated by comparing the two latent minutiae templates against the single reference minutiae template. The texture similarity score ($S_{TT}$) is generated by comparing the latent and reference print texture templates.  The final similarity score $S$ between the latent and the reference print is computed as the weighted sum of $S_{MT,1}$, $S_{MT,2}$ and $S_{TT}$ as below:
\begin{equation}
\label{eq:FinalScore}
S =\lambda_1 S_{MT,1} + \lambda_2 S_{MT,2} +\lambda_3 S_{TT},
\end{equation}
where $\lambda_1$, $\lambda_2$ and $\lambda_3$ are the weights. We empirically determine the values of $\lambda_1$, $\lambda_2$ and $\lambda_3$ to be  1, $1$ and 2, respectively. 

\begin{figure}[t]
	
	\begin{center}
		\includegraphics[scale=0.3]{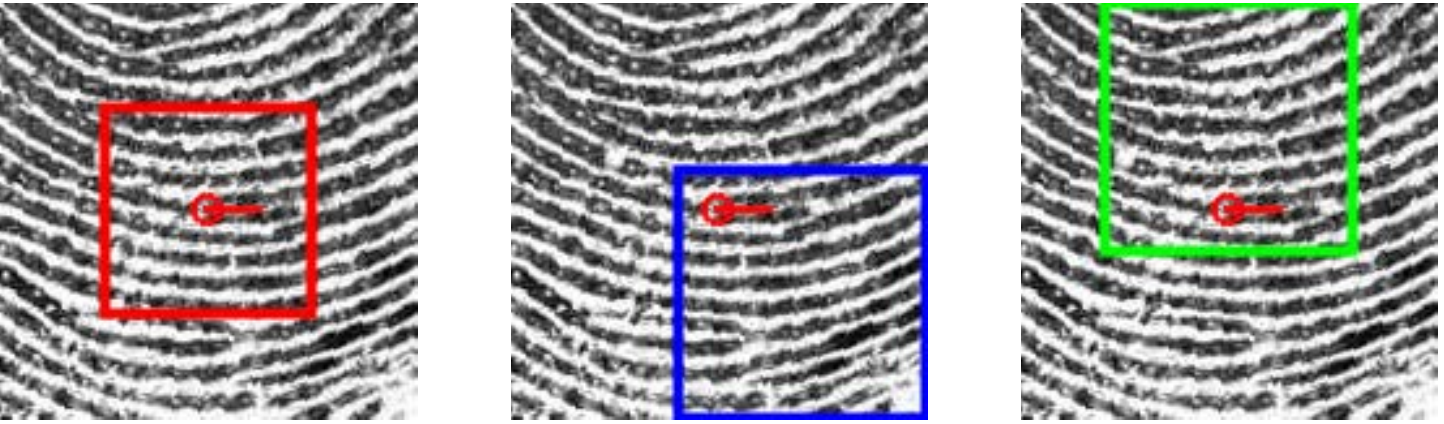}
		
	\end{center}
	\caption{Three selected patch types. The window size of the leftmost is $80\times 80$ pixels. The other two windows are both of size $96\times 96$ pixels. }
	\label{fig:SelectedPatchTypes}
\end{figure}

\begin{figure*}[h]
	
	\begin{center}
		\subfigure[]{
			\includegraphics[width=0.35\linewidth]{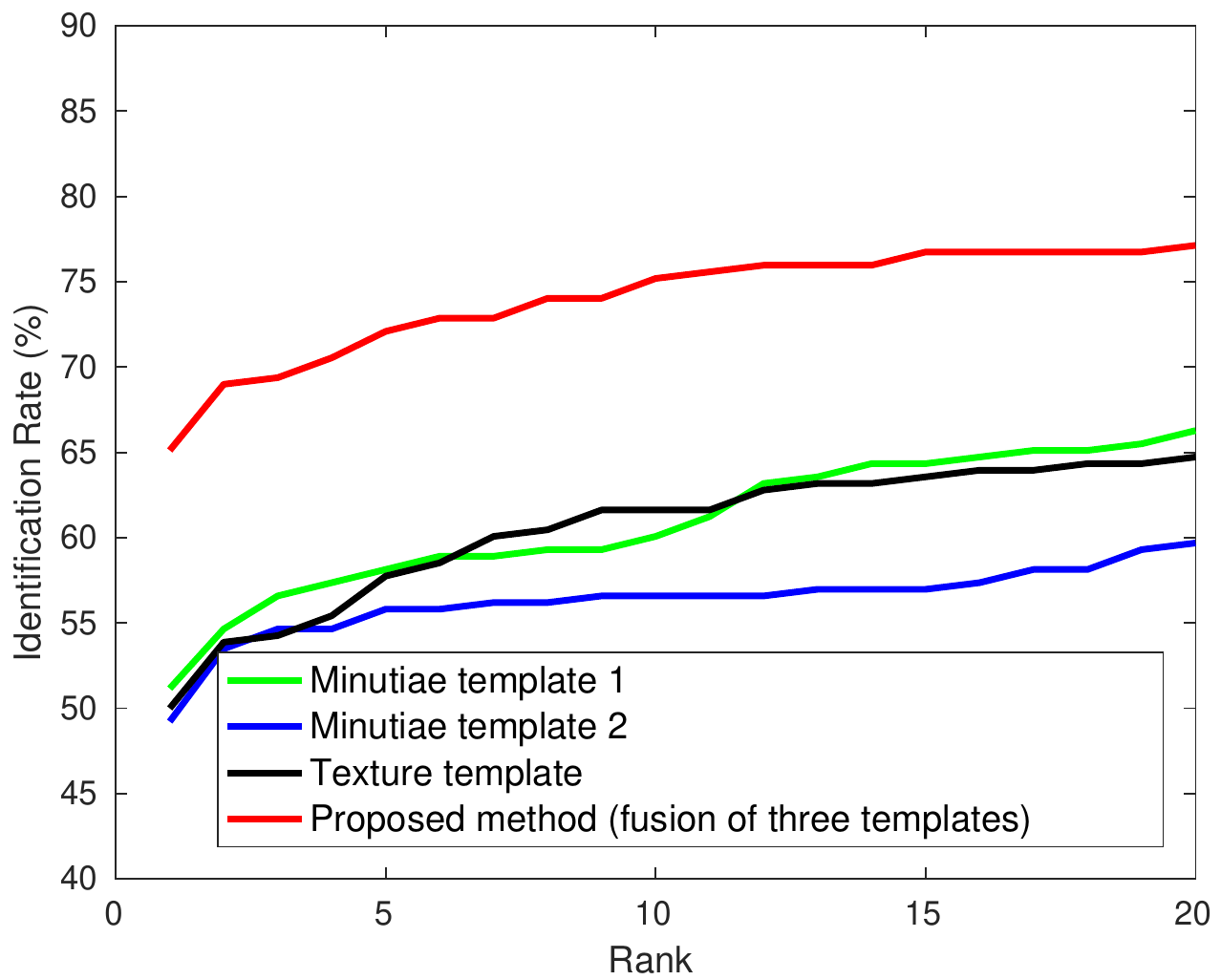}
		}\hspace{0.8cm}
		\subfigure[]{
			\includegraphics[width=0.35\linewidth]{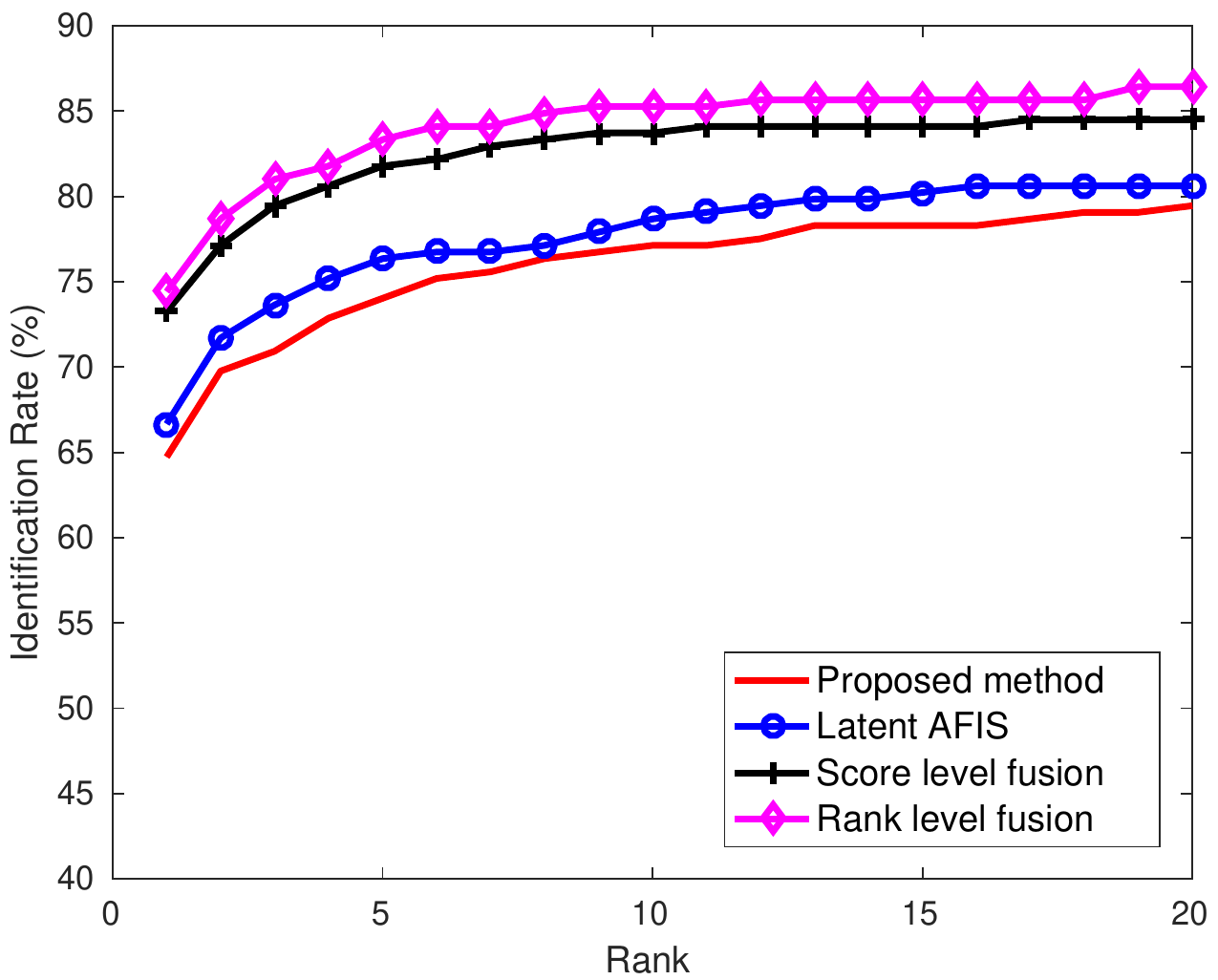}
		}\hspace{0.1cm}
	\end{center}
	\caption{Cumulative Match Characteristic (CMC) curves for NIST SD27 of (a) individual templates (minutia template 1, minutia template 2 and texture template) and their fusion, and (b) comparison of the proposed method with a COTS latent AFIS and score-level and rank-level fusion of the proposed method and COTS latent AFIS.  }
	\label{fig:NIST27}
\end{figure*}

\begin{figure*}[h]
	
	\begin{center}
		\subfigure[]{
			\includegraphics[width=0.35\linewidth]{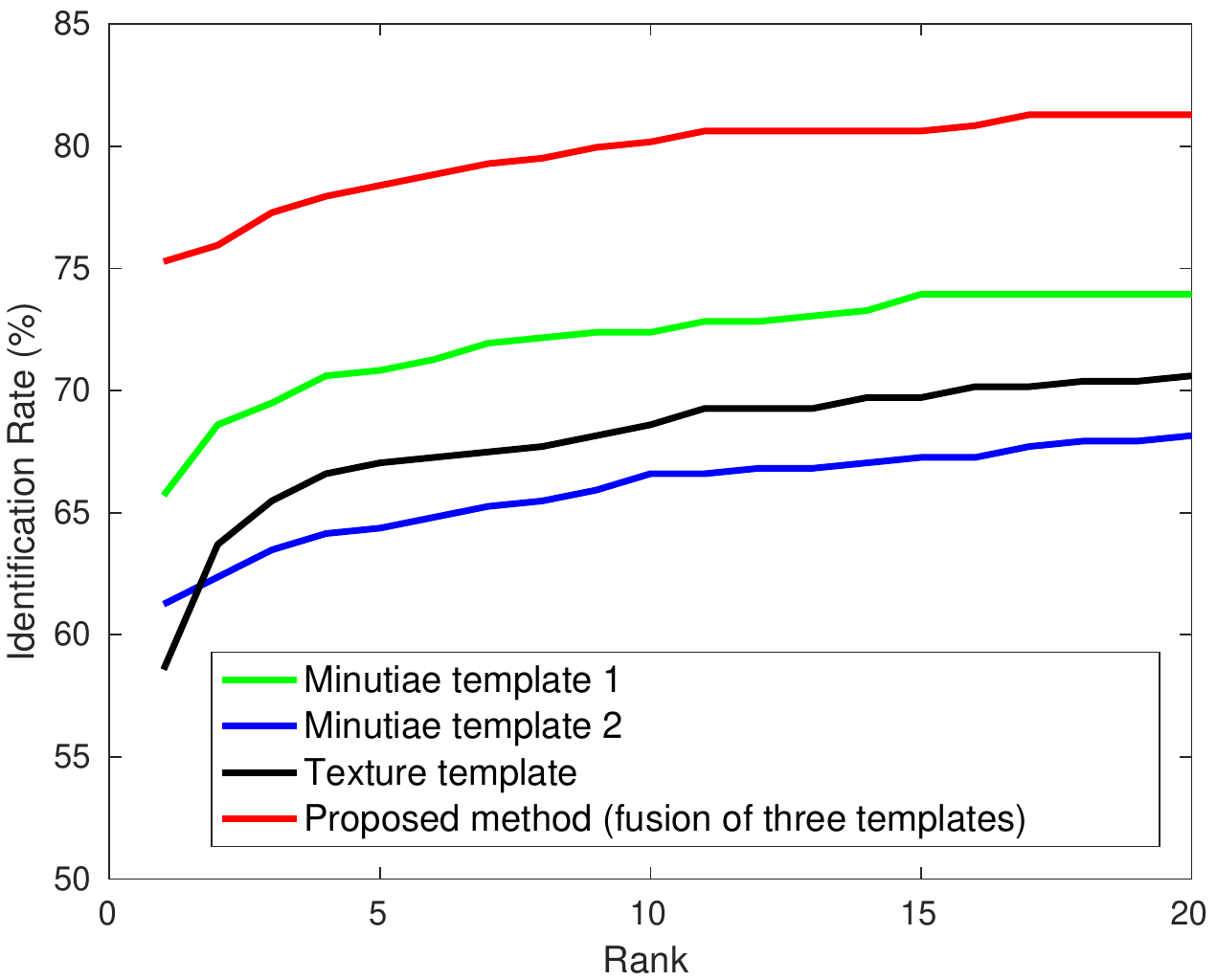}
		}\hspace{0.8cm}
		\subfigure[]{
			\includegraphics[width=0.35\linewidth]{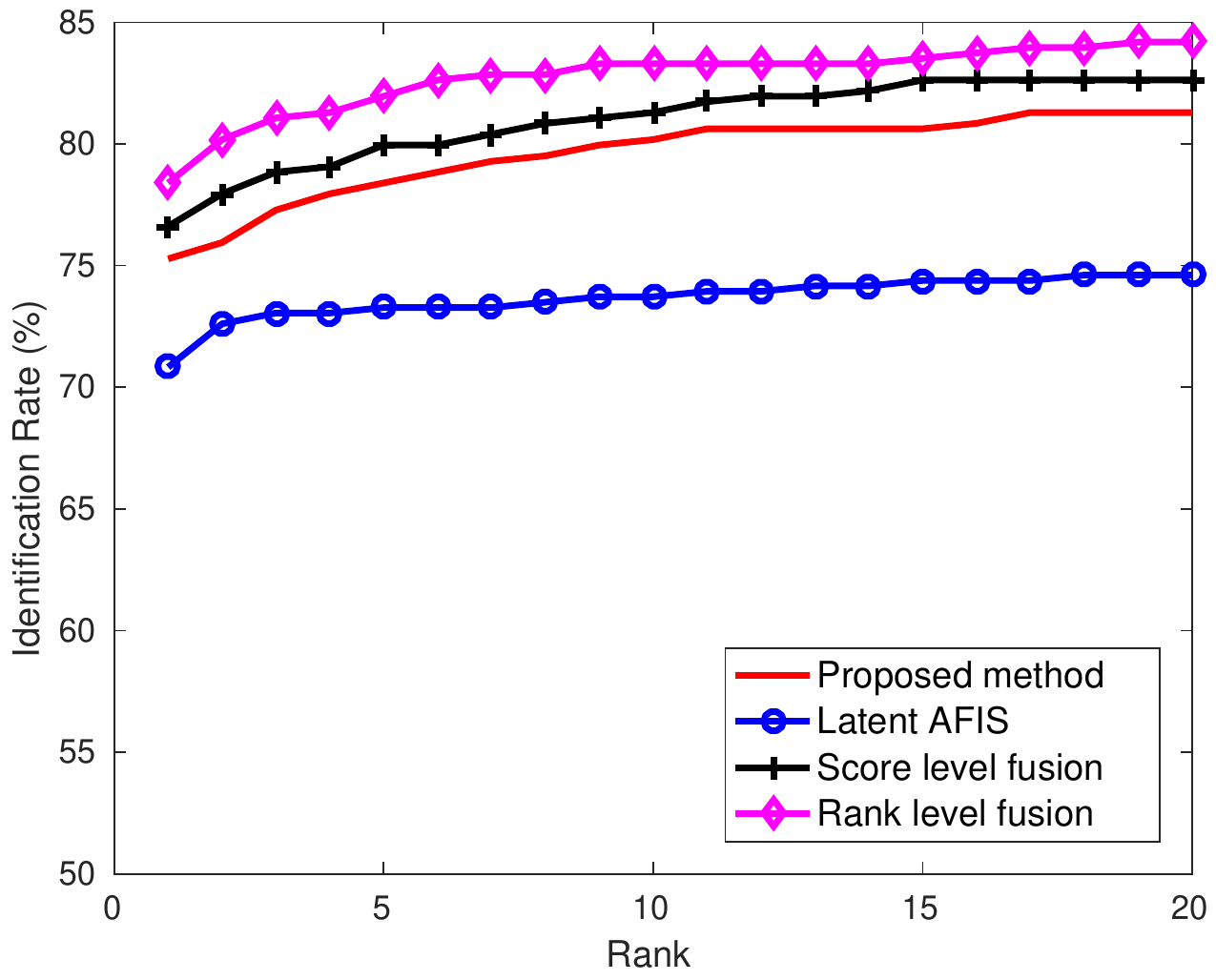}
		}\hspace{0.1cm}
	\end{center}
	\caption{Cumulative Match Characteristic (CMC) curves for WVU DB of (a) individual templates (minutia template 1, minutia template 2 and texture template) and their fusion, and (b) comparison of the proposed method with a COTS latent AFIS and score-level and rank-level fusion of the proposed method and COTS latent AFIS.   }
	\label{fig:WVU}
\end{figure*}

\section{Experimental Results}

\begin{figure}[t]
	
	\begin{center}
		\subfigure[]{
			\includegraphics[width=0.4\linewidth]{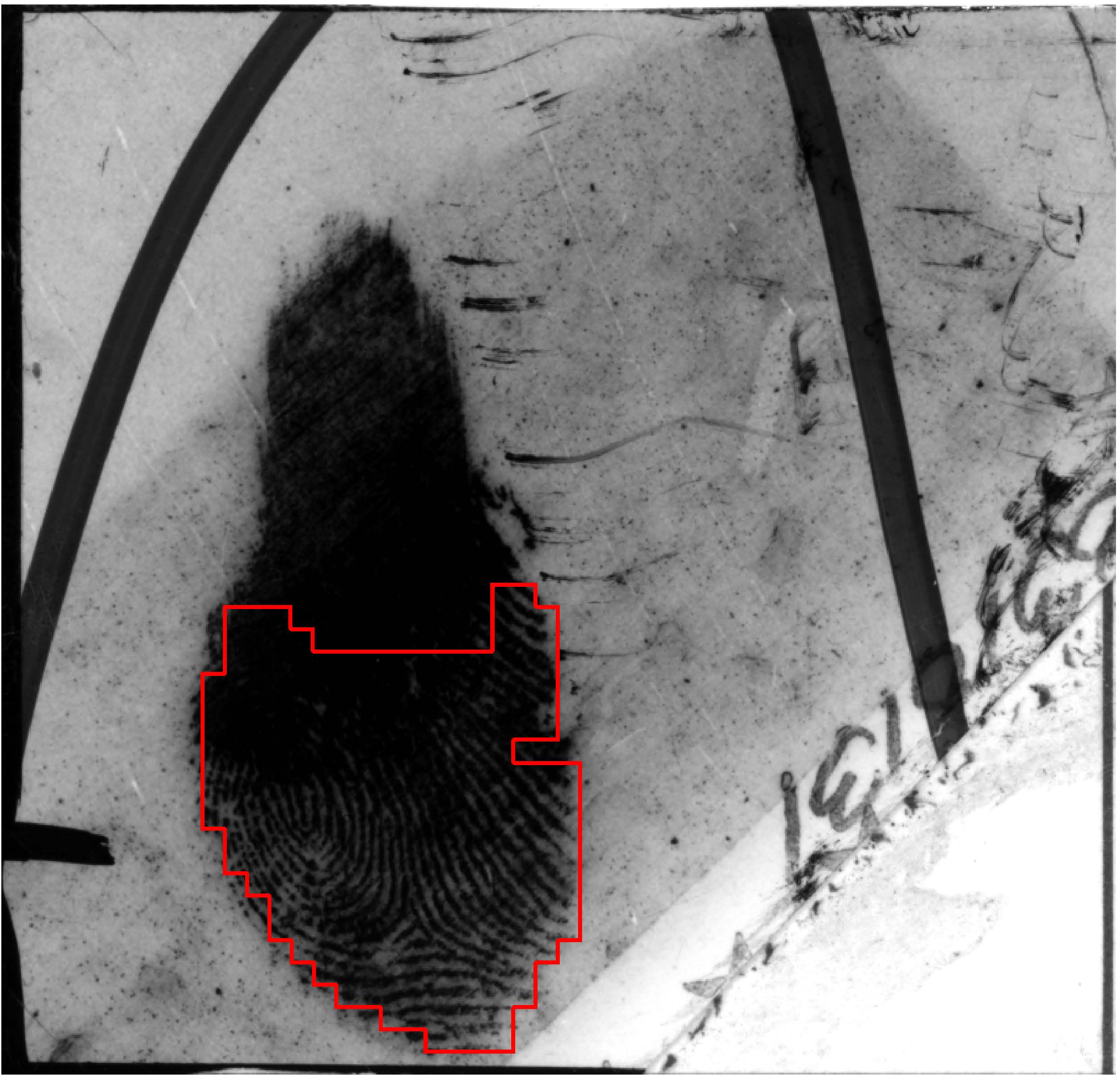}
		}\hspace{0.2cm}
		\subfigure[]{
			\includegraphics[width=0.4\linewidth]{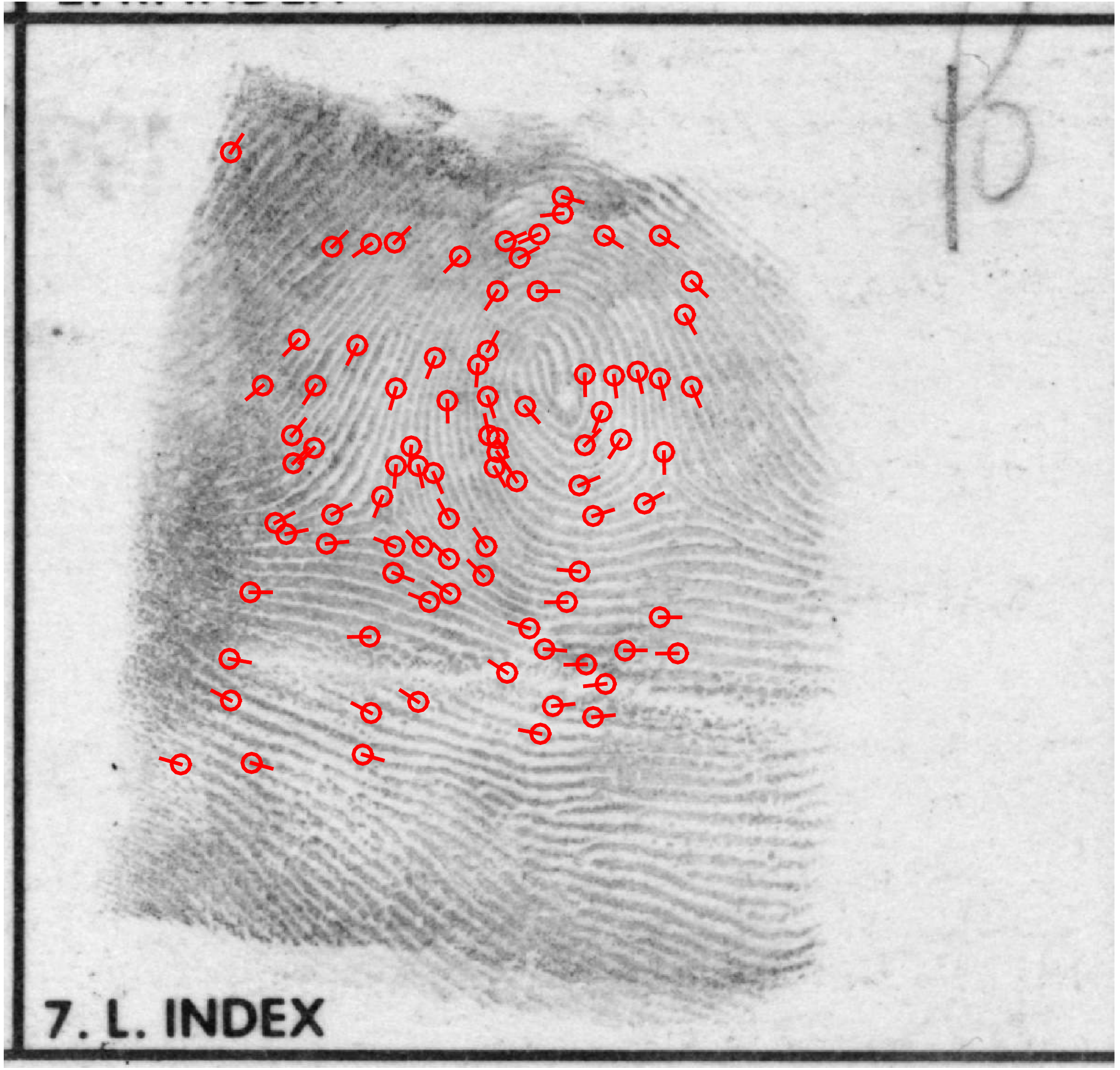}
		}\hspace{0.1cm}
		\subfigure[]{
			\includegraphics[width=0.3\linewidth]{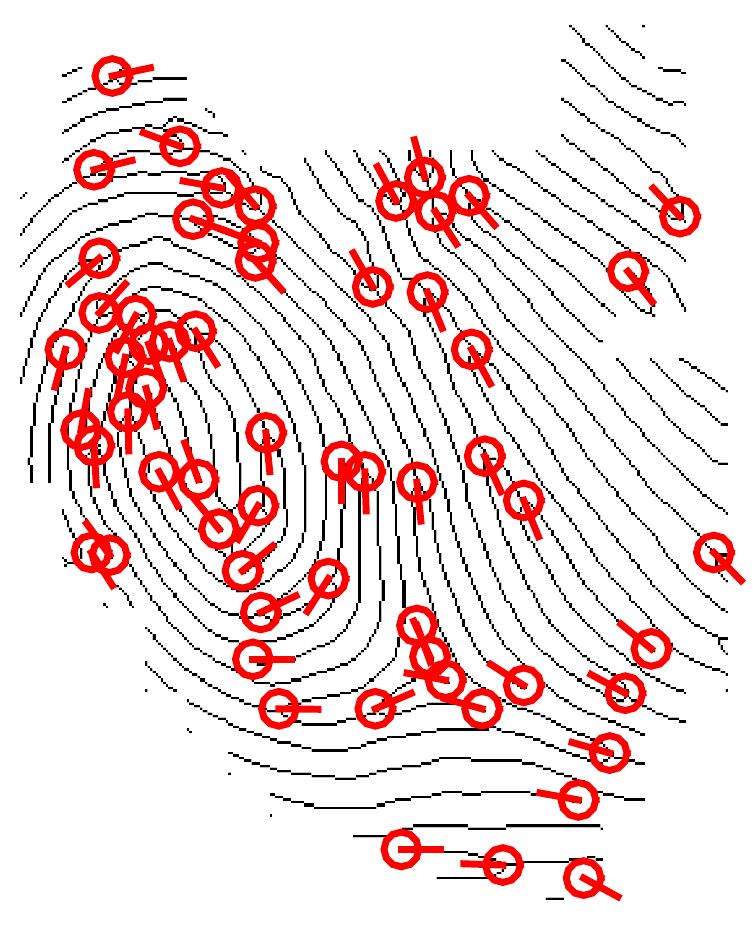}
		}\hspace{0.8cm}
		\subfigure[]{
			\includegraphics[width=0.3\linewidth]{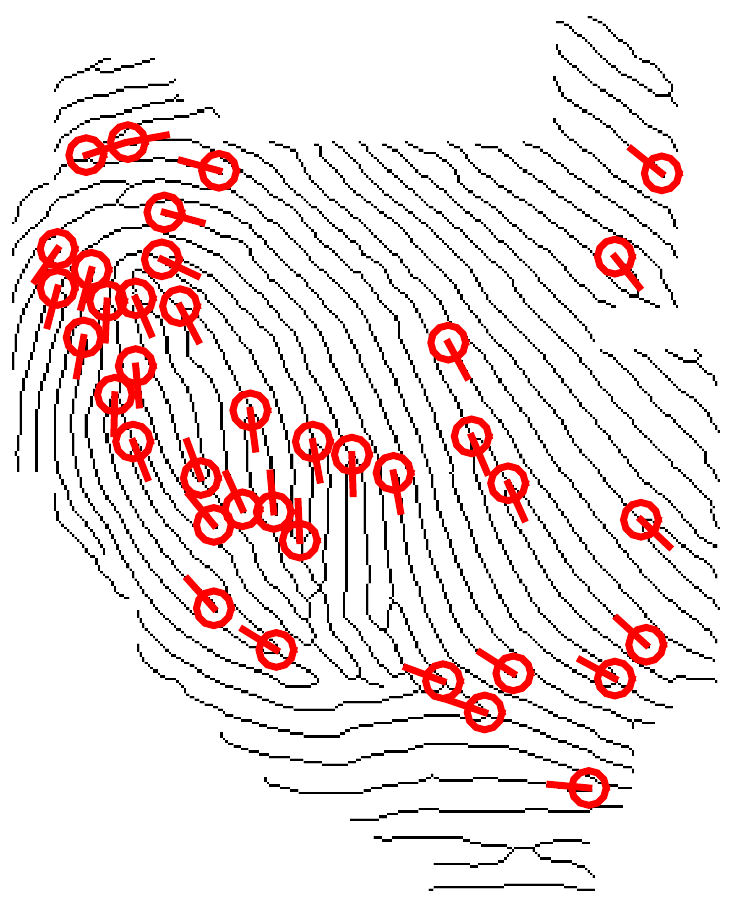}
		}
	\end{center}
	\caption{A latent whose true mate was retrieved at rank-1 by  minutiae template 1 but not by minutiae template 2 (rank-2,457). (a) Input latent with its ROI (G044 from NIST SD27), (b) mated reference print of (a)  with overlaid minutiae, (c)  minutiae set 1 of (a) overlaid on latent skeleton, and (d) minutiae set 2 of (a) overlaid on latent skeleton.}
	\label{fig:Latent_039}
\end{figure}

\begin{figure}[t]
	
	\begin{center}
		\subfigure[]{
			\includegraphics[width=0.4\linewidth]{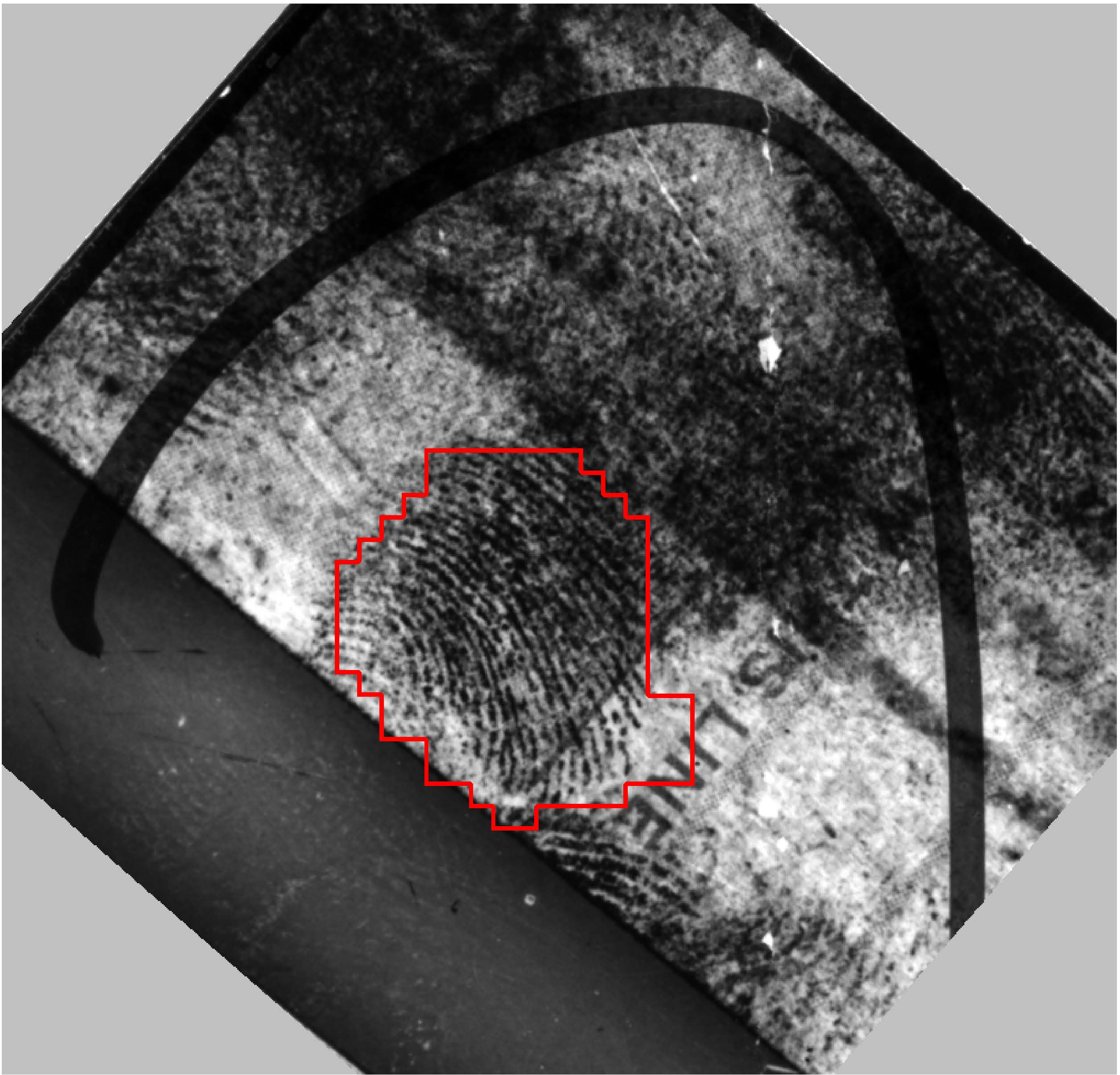}
		}\hspace{0.2cm}
		\subfigure[]{
			\includegraphics[width=0.4\linewidth]{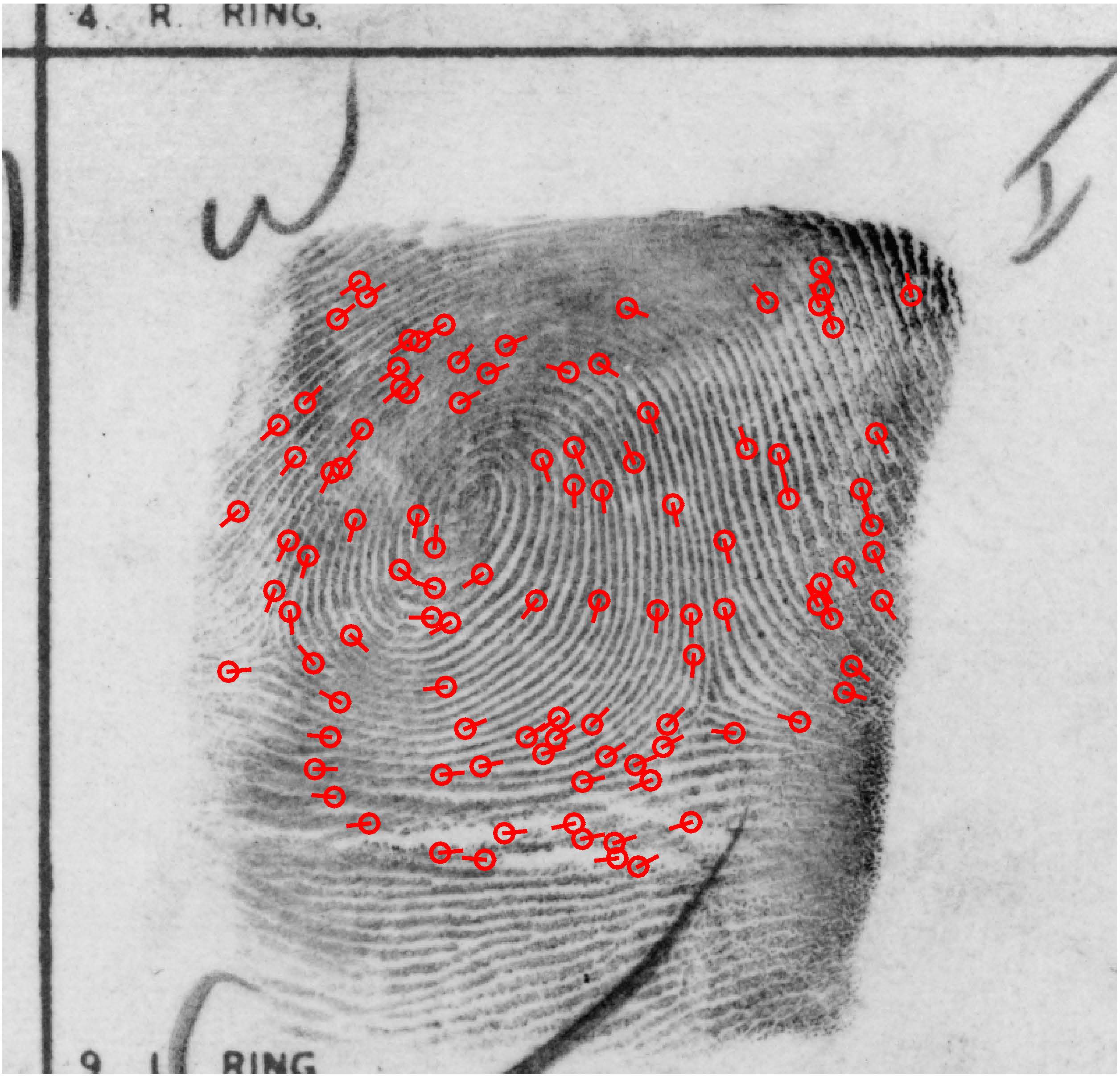}
		}\hspace{0.1cm}
		\subfigure[]{
			\includegraphics[width=0.35\linewidth]{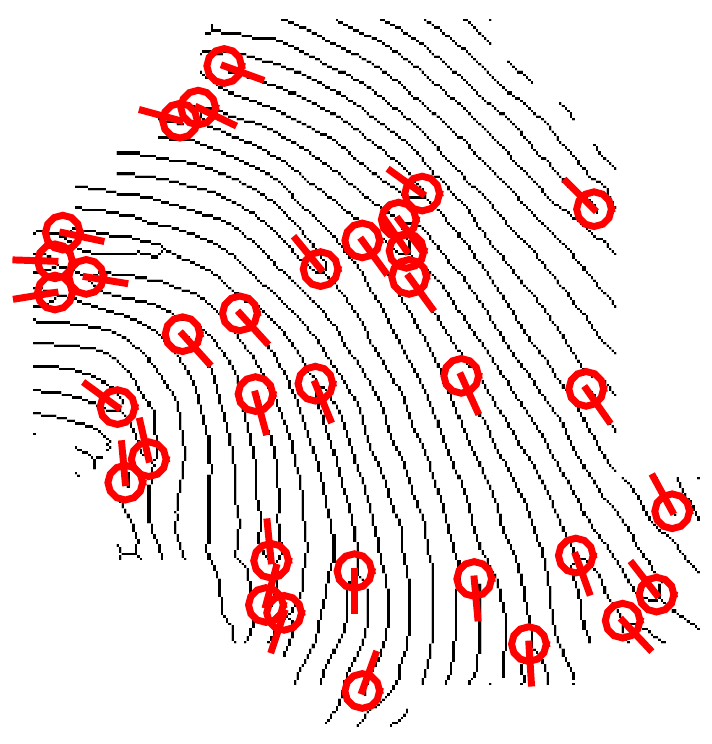}
		}\hspace{0.3cm}
		\subfigure[]{
			\includegraphics[width=0.35\linewidth]{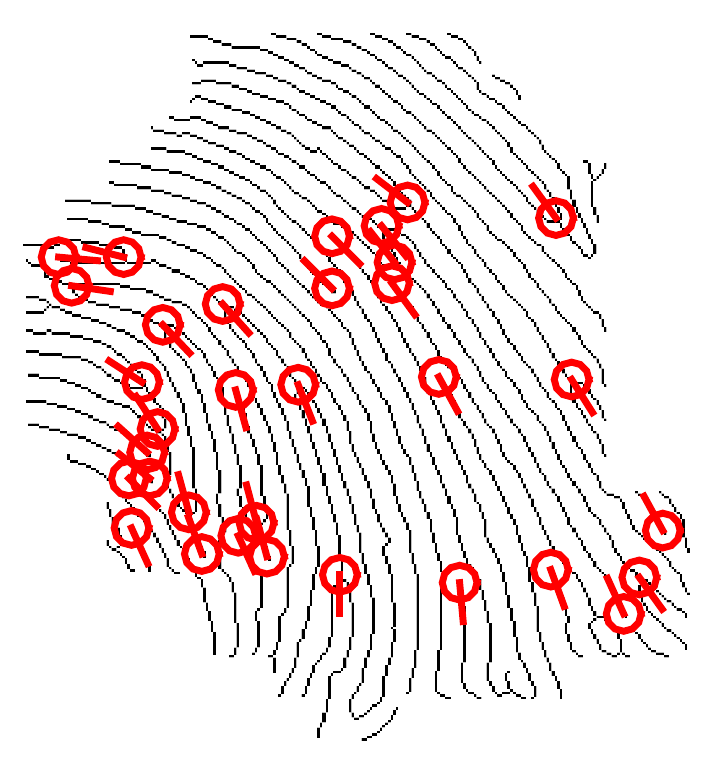}
		}
	\end{center}
	\caption{A latent whose true mate was retrieved at rank-1 by  minutiae template 2 but not by minutiae template 1 (rank-2). (a) Input latent with its ROI (U277 from NIST SD27), (b) mated reference print  with overlaid minutiae, (c)  minutiae set of (a) 1 overlaid on latent skeleton, and (d) minutiae set 2 of (a) overlaid on latent skeleton.}
	\label{fig:Latent_238}
\end{figure}

There is a dearth of latent fingerprint databases available to academic researchers. In this paper, we use two latent databases, NIST SD27 \cite{NISTDB27} and the West Virginia University latent database\footnote{To request WVU latent fingerprint database, contact Dr. Jeremy
	Dawson (Email: Jeremy.Dawson@mail.wvu.edu)} (WVU DB) \cite{WVU} available to us, to evaluate the proposed latent recognition algorithm. The NIST
SD27 contains 258 latent fingerprints with their mated
reference prints. The WVU DB contains 449 latents with their mated reference prints. 
 Note that the NIST SD27 latent database is a collection of latents from the casework of forensics agencies, whereas WVU DB was collected in  a laboratory setting, primarily by students,  at West Virginia University. As such, the characteristics of these two databases are quite different in terms of background noise, ridge clarity, and the number of minutiae. The ridges in some of the latent images in WVU DB are broken apparently  because of dry fingers. See Fig. \ref{fig:FusionWorks} for a comparison of the images in the two databases.

In addition to the mated reference
prints available in these databases, we use additional  reference prints, from NIST SD14 \cite{NISTDB14} and a forensic agency,
to enlarge the reference database to 100,000 for experiments reported here. The larger reference database allows for a challenging latent recognition problem. We follow the protocol used in NIST ELFT-EFS \cite{Indovina2011} \cite{Indovina2012} to evaluate the recognition performance of our system.

The algorithm
was implemented in MATLAB and runs on a server with 12 cores @ 2.50GHz, 256 GB RAM and Linux operating system. Using 24 threads (MATLAB function: \textit{parpool}),  the average template extraction time (all three templates) per latent is  2.7s and the average time for a latent to rolled comparison is 9.2ms on NIST SD27. It is neither fair nor possible to compare our algorithm's compute requirement with COTS.  The COTS latent AFIS has been developed over many years and it is optimized for computing performance. Also, we cannot implement the available SDK on the same multicore environment.


\subsection{Selection of ConvNets for Minutiae Descriptor }

Use of all 14 ConvNets, i.e., 14 patch types in Fig. \ref{fig:PatchTypes}, for minutiae descriptor may not be necessary to achieve the optimal recognition performance. We explore feature selection techniques to determine  a subset of these 14 descriptors that will maintain the latent recognition accuracy. A sequential forward  selection (SFS) \cite{PatternRecognition} of the 14 patch types, using rank-1 accuracy as the criterion on the NIST SD27 database, revealed that  3 out of 14 patch types (Fig. \ref{fig:SelectedPatchTypes})  are adequate without a significant loss in accuracy (75.6\% v. 74.4\%)  yet giving us a significant speed up. 
In the following experiments, we use only these 3 patch types.

\subsection{Performance of Individual Latent Templates}
\begin{figure}[t]
	
	\begin{center}
		\subfigure[]{
			\includegraphics[width=0.45\linewidth]{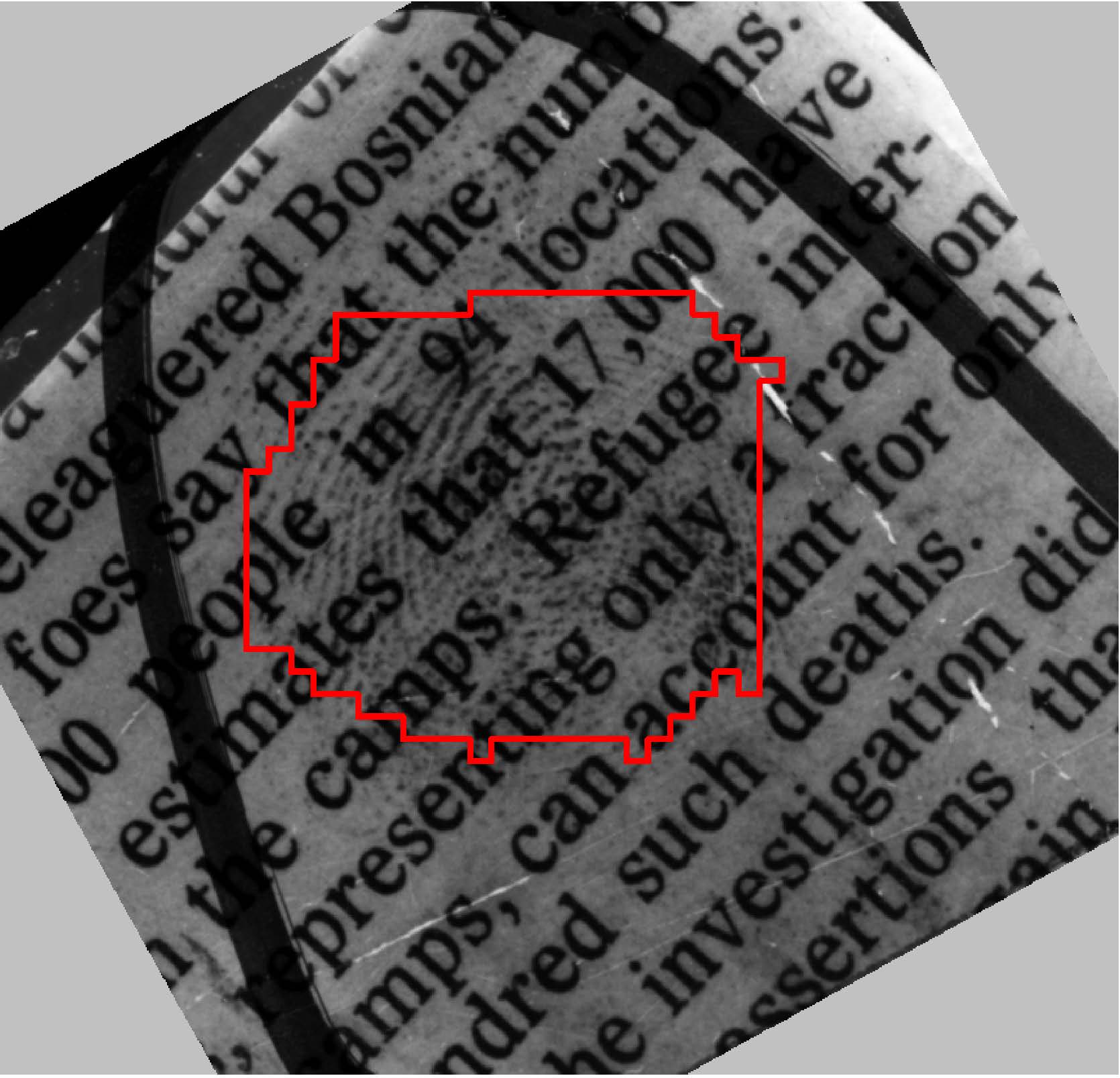}
		}\hspace{0.1cm}
		\subfigure[]{
			\includegraphics[width=0.45\linewidth]{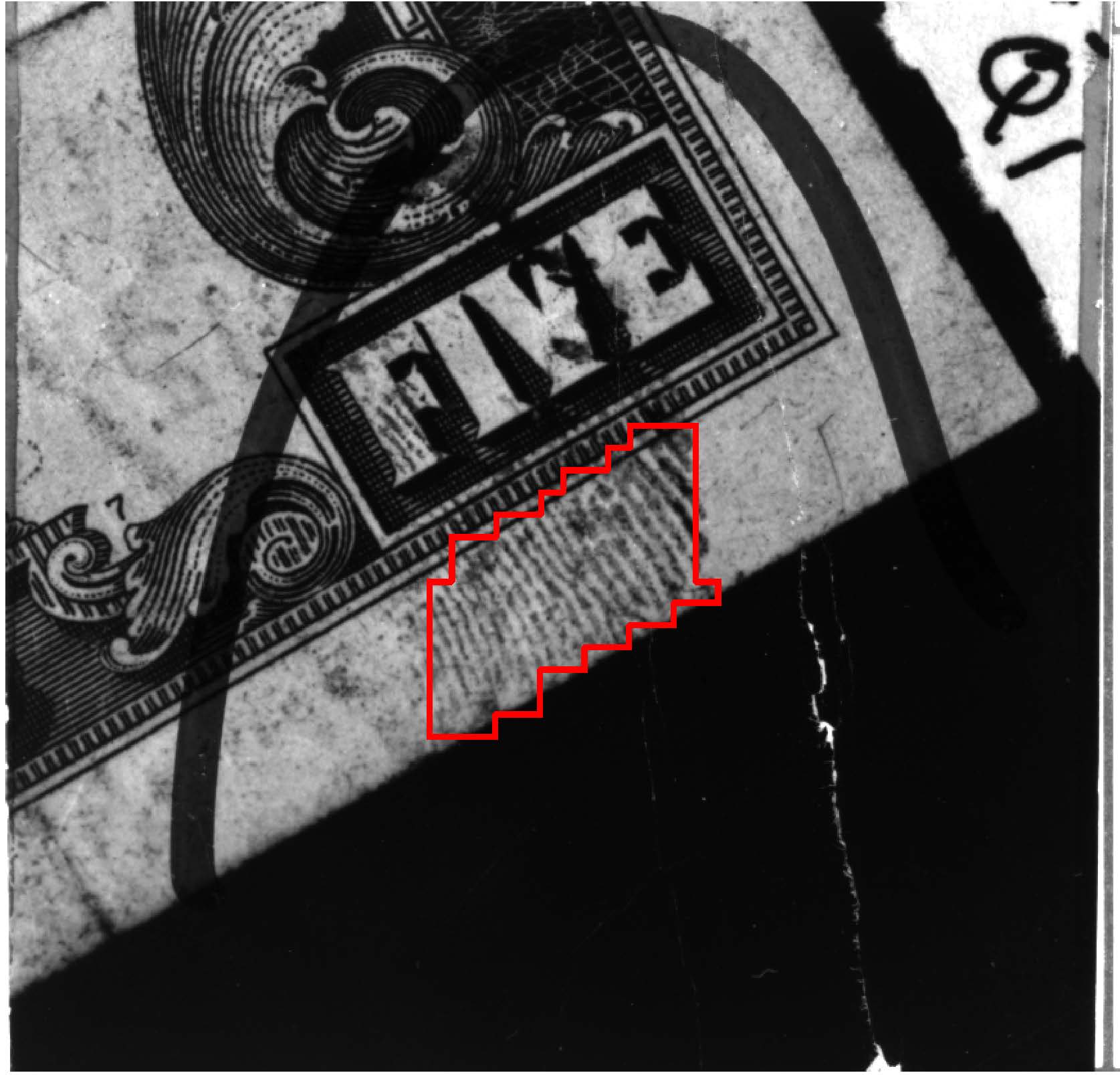}
		}\hspace{0.1cm}
	\end{center}
	\caption{Example latents whose true mates were found at rank-1 by texture template but not by the minutiae templates. Reliable minutiae from these two latents could not be extracted due to (a) poor quality (U276 from NIST SD27) and (b) small friction ridge area (U292 from NIST SD27).}
	\label{fig:TextureWorks}
\end{figure}

Our objective for designing three different templates is to extract complementary information from latents.  
Fig. \ref{fig:NIST27} (a) and Fig. \ref{fig:WVU} (a) compare the Cumulative Match Characteristic (CMC) curves of the three individual templates, namely, minutiae template 1, minutiae template 2 and texture template, on NIST SD27 and WVU DB, respectively. The minutiae template 1 performs significantly better than the minutiae template 2 on both latent databases.  The main reason is that the ridge flow used for generating  minutiae set 1, based on ConvNet,  is more robust than minutiae set 2 extractor, based on ridge flow dictionary. Note that the performance of texture template, which does not utilize any of the true minutiae in latents, is close to the performance of minutiae template 2 on both NIST SD27 and  WVU DB. This can be attributed to the virtual minutiae representation in the texture template and corresponding descriptors extracted by ConvNets.
 Fig. \ref{fig:Latent_039}  shows an example latent whose true mate can be retrieved at rank 1 using minutiae template 1 but not minutiae template 2.  The main reason is that the extracted ridge flow for this latent  is better around the  lower core point for minutiae template 1  than minutia template 2. 
The true mate of the latent shown in Fig. \ref{fig:Latent_238} (a) can be retrieved at rank 1 using minutiae template 2 but not minutiae template 1 even though their skeletons look similar.   Fig. \ref{fig:TextureWorks} shows two latent examples which  lack reliable minutiae but the texture template is able to find their true mates at rank 1. 

We also evaluate fusion of different subsets of the three templates. The fusion of any two templates using the weights in Eq. (\ref{eq:FinalScore}) performs better than any single template, and the performance can be further improved by fusing all three templates. This demonstrates that the three templates proposed here contain complementary information for latent recognition. Most significantly, the texture template, in conjunction with the two minutiae templates boosts the overall recognition performance (from 58.5\% to 64.7\% rank-1 accuracy on NIST SD27 and from 70.6\% to 75.3\% on WVU DB). 

 \begin{figure}[t]
	
	\begin{center}
		\subfigure[]{
			\includegraphics[width=0.4\linewidth]{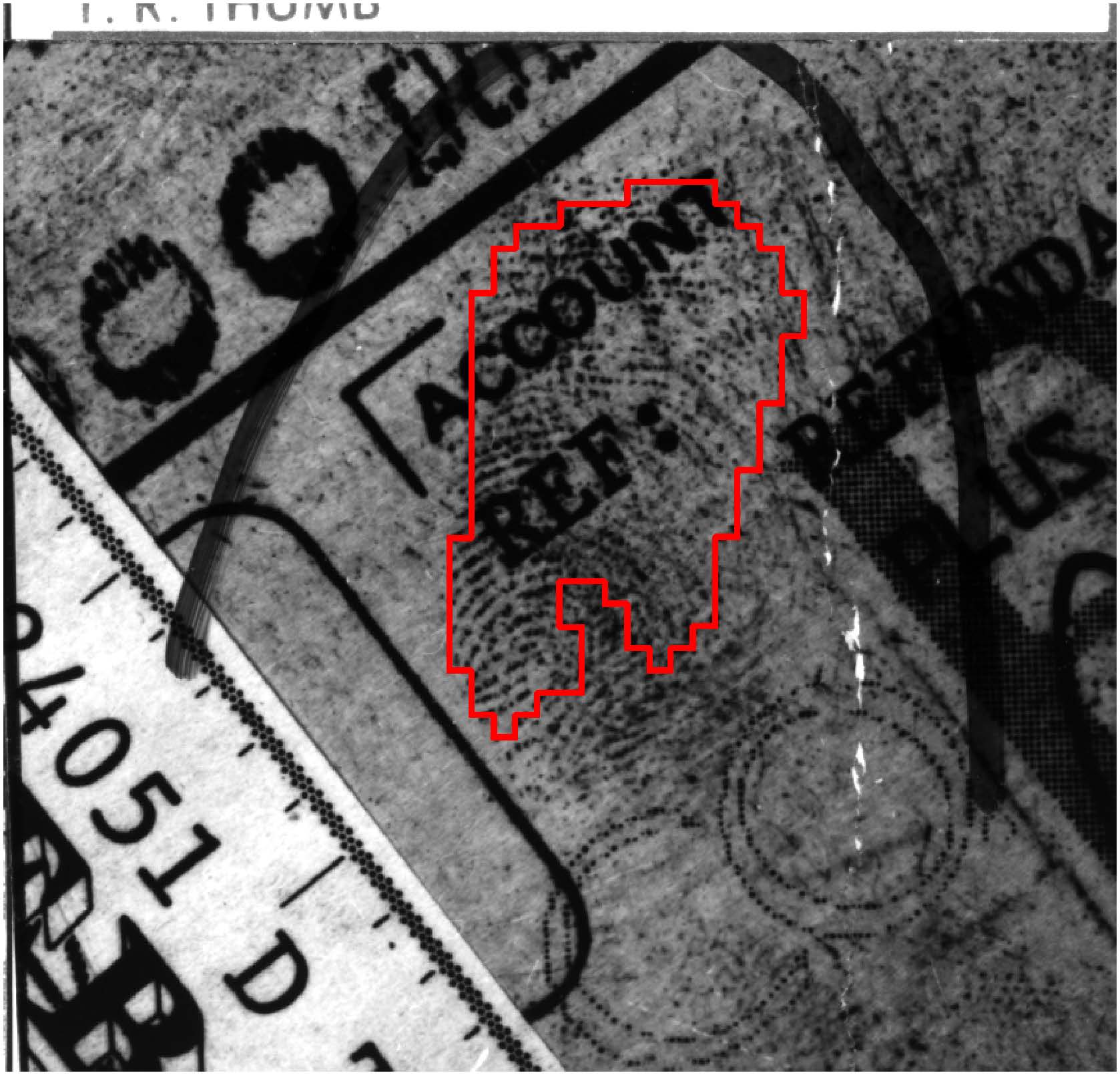}
		}\hspace{0.1cm}
		\subfigure[]{
			\includegraphics[width=0.4\linewidth]{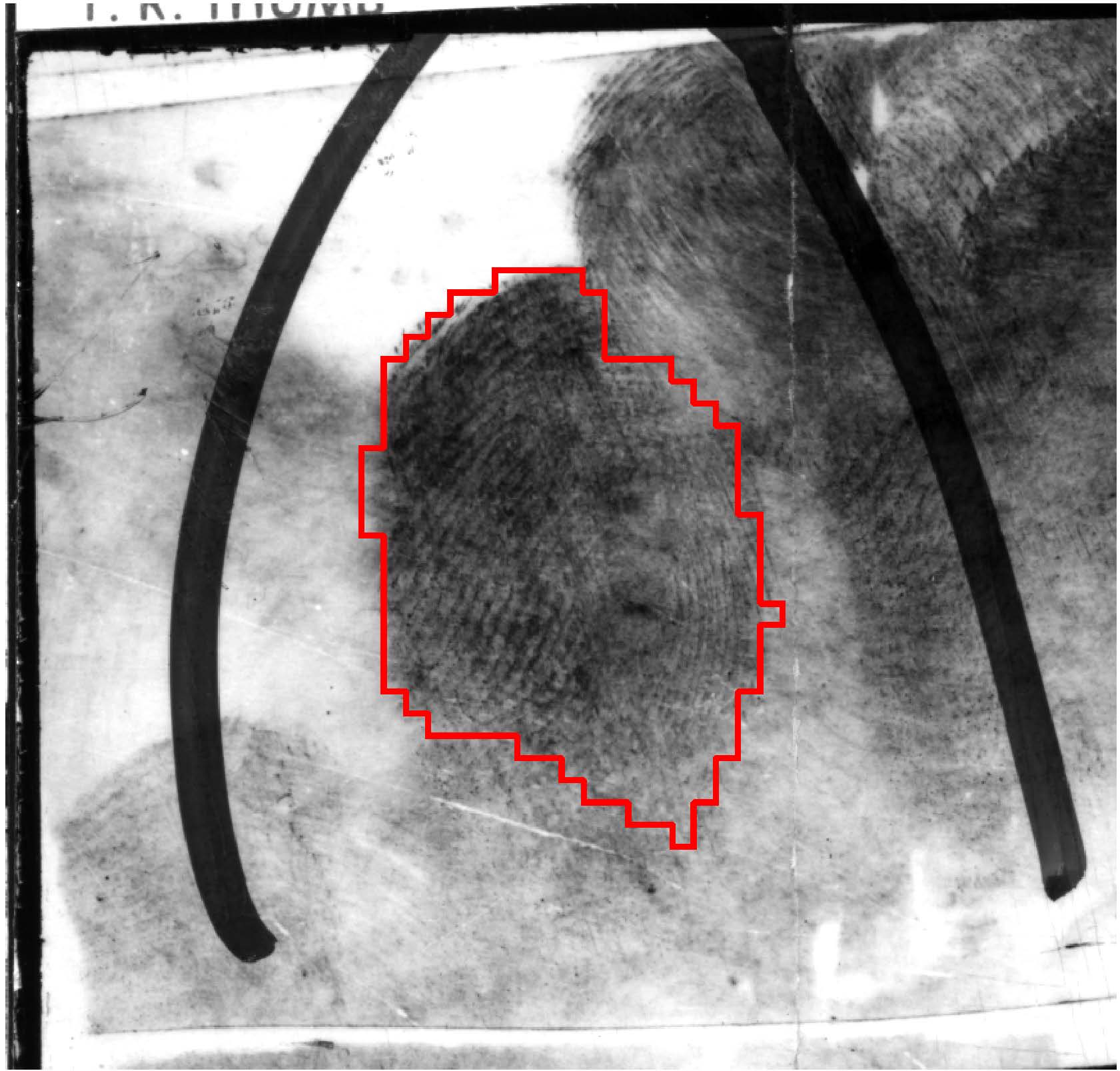}
		}\hspace{0.1cm}
		\subfigure[]{
			\includegraphics[width=0.3\linewidth]{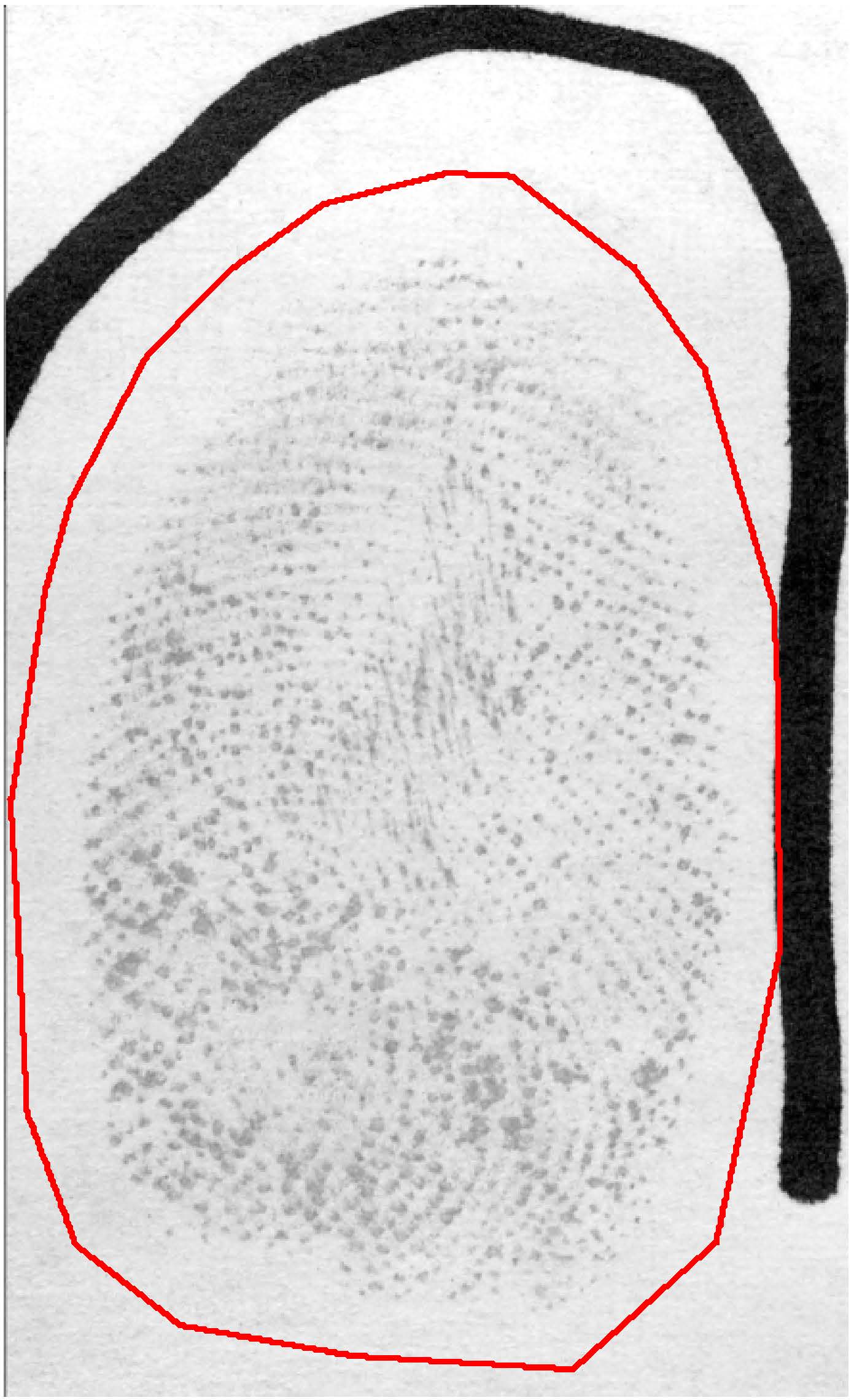}
		}\hspace{0.3cm}
		\subfigure[]{
			\includegraphics[width=0.35\linewidth]{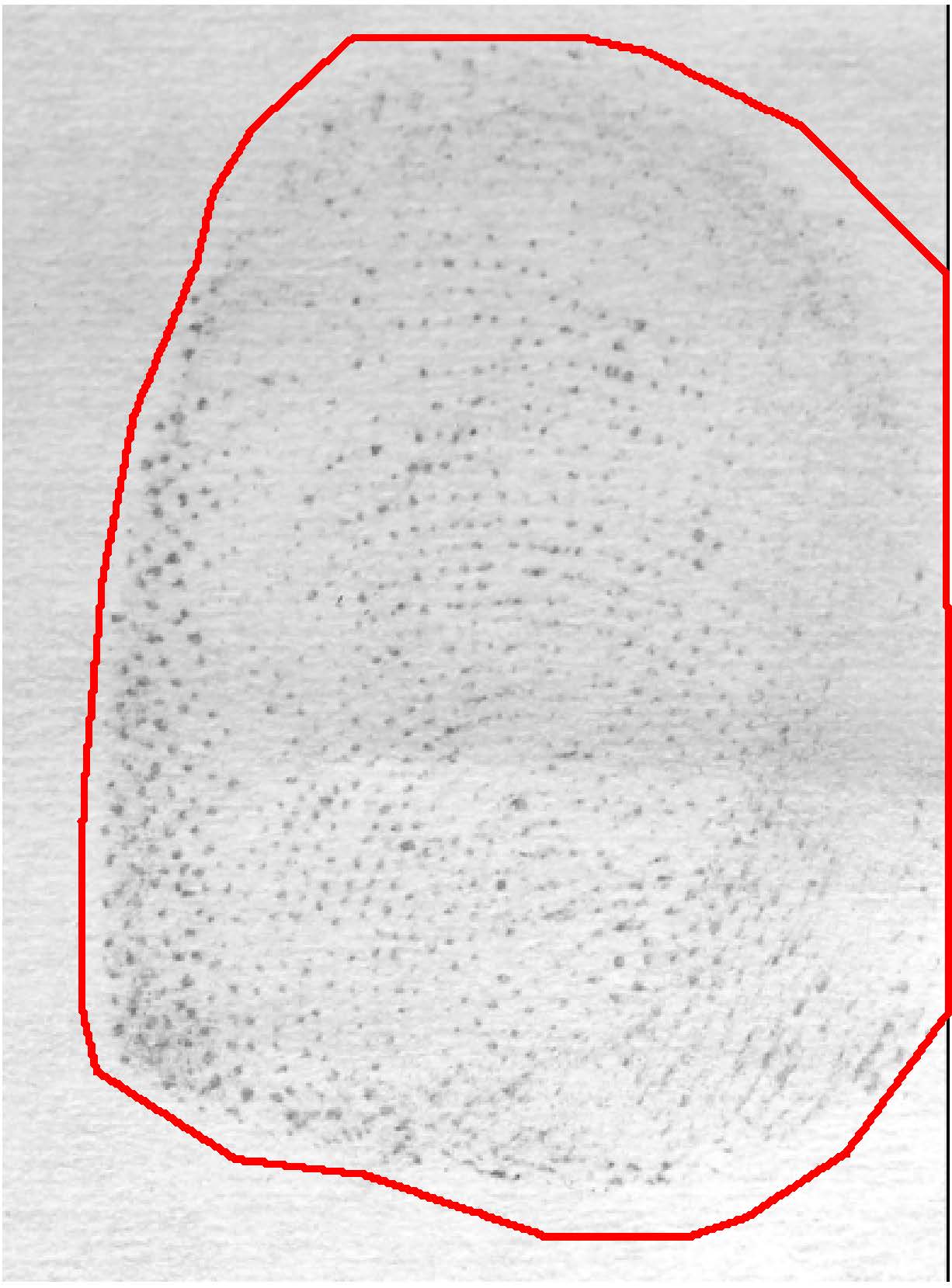}
		}
	\end{center}
	\caption{Example of latent images which are correctly identified at rank-1 by the proposed method but not by a leading COTS latent AFIS.  The retrieval rank of the true mate of (a) by the latent AFIS is 931, but for latents in (b), (c) and (d), their true mates could not be found because the comparison score was zero.  Latents in (a) and (b) are from NIST SD27 whereas latents in (c) and (d) are from WVU DB. }
	\label{fig:FusionWorks}
\end{figure}

\subsection{Benchmarking against COTS Latent AFIS}

We benchmark the proposed latent recognition algorithm against one of the best COTS latent AFIS\footnote{The latent AFIS used here is one of top-three performers in the NIST ELFT-EFS evaluations \cite{Indovina2011} \cite{Indovina2012}.  Because of our non-disclosure agreement with the vendor, we cannot disclose the name. }  as determined in NIST evaluations.  The input to the latent AFIS are cropped latents using the same ROI as input to the proposed algorithm. 
While the COTS latent AFIS performs slightly better than the proposed algorithm (Rank-1 accuracy of 66.7\% for COTS latent AFIS vs. 64.7\%  for the proposed algorithm ) on NIST SD27,  the proposed method outperforms the COTS latent AFIS on WVU DB  (Rank-1 accuracy of 75.3\% vs. 70.8\%). See Figs.  \ref{fig:NIST27} (b) and \ref{fig:WVU} (b). The overall recognition performance can be further improved by a fusion of the proposed algorithms and COTS latent AFIS. Two fusion strategies, namely score-level fusion (with equal weights) and rank-level fusion (two top 100 candidates lists are fused using Borda count \cite{Ross2006})  were implemented.
Score level fusion of the COTS and the proposed algorithm results in significantly higher rank-1 accuracies, i.e., 73.3\% on NIST SD27 and  76.6\% on WVU DB. For NIST SD27 with a total of 258 latents, the score-level fusion leads to an additional 17 latents whose mates are now retrieved at rank-1 compared to the COTS latent AFIS alone. Rank-level fusion results in even better performance (Rank-1 accuracies of 74.4\% on NIST SD27 and  78.4\% on WVU DB ).

Note that rank-level fusion is preferred over score-level fusion when, for proprietary reasons, a vendor may not be willing to reveal the comparison scores. The CMC curves are shown in Figs. \ref{fig:NIST27} and \ref{fig:WVU}. Fig. \ref{fig:FusionWorks} shows example latents whose true mates can be correctly retrieved  at rank-1 by the proposed method, but the COTS latent AFIS was not successful. Although the two example latents from WVU DB (Figs. \ref{fig:FusionWorks} (c) and (d)) have large friction ridge area, the latent AFIS outputs comparison scores of 0 between the latents and their  mates. Apparently, the latent AFIS could not extract sufficient number of reliable minutiae in the latents where the ridges are broken. The proposed algorithm with its use of two different ridge flow estimation algorithms and dictionary-based and Gabor filtering-based enhancement, is able to obtain high quality ridge structures  and sufficient number of minutiae.

To compare the proposed ConvNet-based minutiae descriptor with MCC descriptor \cite{Cappelli2010PAMI} which is a popular  minutiae descriptor  for reference prints, we replace the ConvNet-based descriptor in latent minutiae template 1  and reference print minutiae template by MCC descriptor. The rank-1 accuracies on NIST SD27 and WVU DB by comparing modified minutiae template 1 of latents against modified minutiae templates of 100K reference prints are only 21.3\% and 35.2\%, respectively. These accuracies are far lower than the accuracies of the proposed minutiae template 1 with learned descriptors based on ConvNet (rank-1 accuracies of 51.2\% and 65.7\% on NIST SD27 and WVU DB, respectively).

We  also compare the proposed latent recognition algorithm with Paulino et al.'s algorithm \cite{PaulinoTIFS2013},
 which uses manually marked minutiae and MCC descriptor. The rank-1 identification rates of the proposed method are about 20\% and 32\% higher than those reported in Paulino et al. \cite{PaulinoTIFS2013} on NIST SD27 and WVU DB, respectively.


\section{Conclusions and Future Work}

Latent fingerprints constitute one of the most important and widely used sources of forensic evidence  in forensic investigations. Despite this, efforts to design and build accurate, robust, and fully automated latent fingerprint recognition systems have been limited. Only a handful of commercial companies are able to provide large-scale latent SDKs, but even they  require significant time and effort of latent examiners  in finding the true mate or a ``hit"  of a query latent. To our knowledge, open source literature does not contain any automated latent recognition method. The  latent recognition problem is difficult due to poor ridge quality, severe background noise, small friction ridge area, and image distortion  encountered in latent images.  

We present an automated latent fingerprint recognition algorithm and benchmark its performance against a leading COTS latent AFIS.  The contributions of this paper are as follows:
\begin{enumerate}
	\item Three latent templates, namely, two minutiae templates and one texture template, are utilized.  These templates extract complementary information from latents; 
	\item A total of 14 patch types are investigated for minutiae descriptors  that are learned via a ConvNet. A systematic feature selection method shows that only 3 out of 14 patch types are needed to maintain the overall recognition accuracy at a significant savings in computation. 
	\item Second-order and third-order graph based minutiae correspondence algorithms are proposed for establishing minutiae correspondences. 
	\item Experimental results show that the proposed method performs significantly better than published algorithms on two benchmark databases (NIST SD27 and WVU latent DB) against 100K rolled prints. Further,  our algorithm is competitive and complementary to a leading COTS  latent AFIS.   Indeed, a fusion of the proposed method and COTS latent AFIS leads to a boost in rank-1 recognition accuracy for both the benchmark latent databases. 
\end{enumerate}

Our algorithm for latent recognition can be further improved  as follows. 
\begin{enumerate}
\item  ConvNet architectures, e.g., GoogeLeNet \cite{GoogLeNet}, should be considered to improve the recognition effectiveness. 
\item Exploring the use of additional latent features, such as ridge count and singular points, to further boost the recognition performance. 
\item Filtering strategies through a cascaded network of recognition engines should be studied to improve the system scalability for recognition against  large scale reference set.
\item Acquiring a large collection of latents to train the ConvNet.
\item  Improving the speed of feature extraction and comparison.

\end{enumerate}
\ifCLASSOPTIONcaptionsoff
  \newpage
\fi



%
\input{reference.bbl}

\vfill
\begin{IEEEbiography}[{\includegraphics[width=1in, height=1.25in,clip,keepaspectratio]{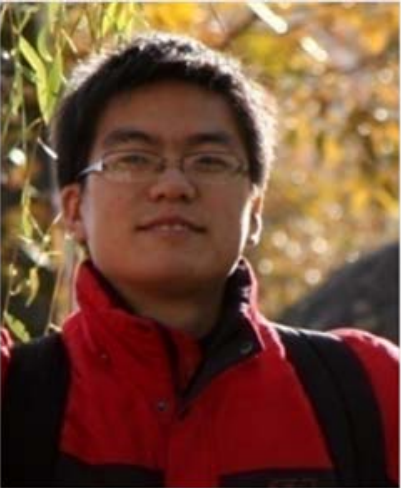}}]{Kai Cao}
 received the Ph.D. degree from the Key Laboratory of Complex
Systems and Intelligence Science, Institute of Automation, Chinese
Academy of Sciences, Beijing, China, in 2010. He is currently a Post
Doctoral Fellow in the Department of Computer Science \&
Engineering, Michigan State University. He was
affiliated with Xidian University as an Associate Professor.
His research interests include biometric recognition, image
processing and machine learning.
\end{IEEEbiography}

\vfill
\begin{IEEEbiography}[{\includegraphics[width=1in, height=1.25in,clip,keepaspectratio]{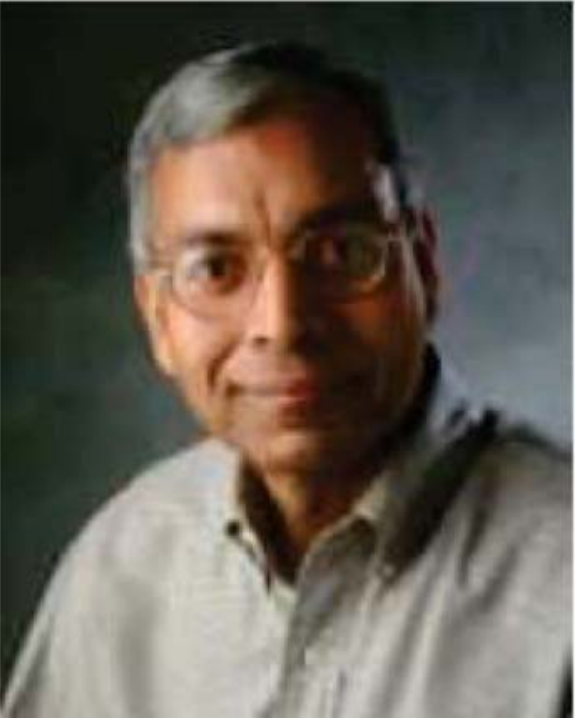}}]{Anil K. Jain}
is a University distinguished professor
in the Department of Computer Science and
Engineering at Michigan State University. His
research interests include pattern recognition
and biometric authentication. He served as the
editor-in-chief of the IEEE Transactions on Pattern
Analysis and Machine Intelligence and was a member of the United
States Defense Science Board. He has received
Fulbright, Guggenheim, Alexander von Humboldt, and IAPR King
Sun Fu awards. He is a member of the National Academy of Engineering
and foreign fellow of the Indian National Academy of Engineering.
\end{IEEEbiography}

\end{document}